\documentclass[10pt,twocolumn,letterpaper]{article}

\usepackage{cvpr}              
\usepackage{booktabs}
\usepackage{multirow}
\usepackage[dvipsnames,table,xcdraw]{xcolor}
\usepackage{makecell}
\usepackage{xspace}
\usepackage{url}
\usepackage{booktabs}
\usepackage{amssymb}
\usepackage{bbding}
\usepackage{pifont}
\usepackage{wasysym}
\usepackage{utfsym}
\usepackage{fontawesome}
\usepackage{indentfirst}
\usepackage{bm}

\usepackage [utf8]{inputenc}
\usepackage{enumitem}

\definecolor{cvprblue}{rgb}{0.21,0.49,0.74}
\usepackage[pagebackref,breaklinks,colorlinks,citecolor=cvprblue]{hyperref}

\def\paperID{1369} 
\def\confName{CVPR}
\def\confYear{2024}
\title{MADTP: Multimodal Alignment-Guided Dynamic Token Pruning for Accelerating Vision-Language Transformer}

\author{Jianjian Cao$^1$ \quad
Peng Ye$^1$ \quad
Shengze Li$^1$ \quad
Chong Yu$^2$ \quad
Yansong Tang$^3$ \quad
\\
Jiwen Lu$^3$ \quad
Tao Chen$^1$\footnotemark[2]\\
$^1$School of Information Science and Technology, Fudan University\\
$^2$Academy for Engineering and Technology, Fudan University\\
$^3$Tsinghua-Berkeley Shenzhen Institute, Tsinghua University\\
{\tt\small jjcao22@m.fudan.edu.cn, eetchen@fudan.edu.cn}
}

\begin{document}
\maketitle
\renewcommand{\thefootnote}{\fnsymbol{footnote}}
\footnotetext[2]{Corresponding authors.}
\vspace{-5mm}
\begin{abstract}
Vision-Language Transformers (VLTs) have shown great success recently, but are meanwhile accompanied by heavy computation costs, where a major reason can be attributed to the large number of visual and language tokens.
Existing token pruning research for compressing VLTs mainly follows a single-modality-based scheme yet ignores the critical role of aligning different modalities for guiding the token pruning process, causing the important tokens for one modality to be falsely pruned in another modality branch. Meanwhile, existing VLT pruning works also lack the flexibility to dynamically compress each layer based on different input samples.
To this end, we propose a novel framework named \textbf{M}ultimodal \textbf{A}lignment-Guided \textbf{D}ynamic \textbf{T}oken \textbf{P}runing (\textbf{MADTP}) for accelerating various VLTs. 
Specifically, we first introduce a well-designed Multi-modality Alignment Guidance (MAG) module that can align features of the same semantic concept from different modalities, to ensure the pruned tokens are less important for all modalities. We further design a novel Dynamic Token Pruning (DTP) module, which can adaptively adjust the token compression ratio in each layer based on different input instances. Extensive experiments on various benchmarks demonstrate that MADTP significantly reduces the computational complexity of kinds of multimodal models while preserving competitive performance. Notably, when applied to the BLIP model in the NLVR2 dataset, MADTP can reduce the GFLOPs by 80\% with less than 4\% performance degradation. The code is available at \href{https://github.com/double125/MADTP}{https://github.com/double125/MADTP}.
\end{abstract}    
\vspace{-5mm}
\section{Introduction}
\label{sec:intro}
Vision-Language Transformers (VLTs) have taken multimodal learning domain by storm due to their superior performance on various multimodal tasks, including Visual Reasoning~\cite{Johnson_2017_CVPR}, Image Captioning~\cite{Imagecaption_2014}, Image-Text Retrieval~\cite{ITretriveal_2015}, and Visual Question Answering (VQA)~\cite{VQA_2015}. However, these models~\cite{CLIP,kim2021vilt,li2022blip,li2023blip,gpt4}, such as CLIP~\cite{CLIP} and BLIP~\cite{li2022blip}, inevitably suffer from expensive computational costs due to their complex architecture, large parameters, and numerous tokens, which restrict their real-world applications and deployments.

\begin{figure}[t]
  \centering
  \includegraphics[width=\linewidth]{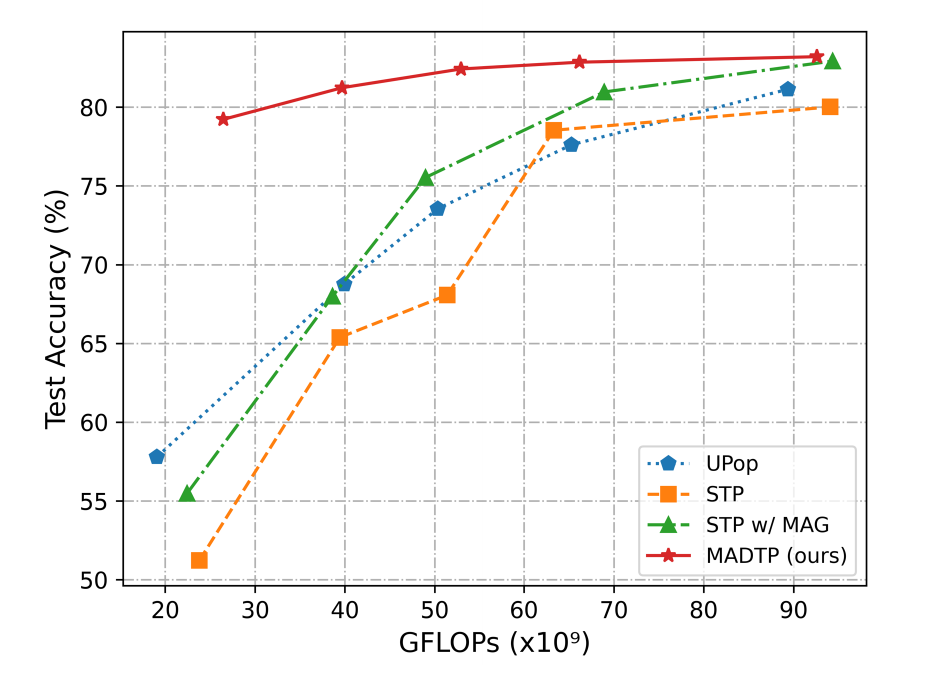}
  \vspace{-9mm}
  \caption{Comparison between our MADTP and other compression methods for the BLIP model tested on the NLVR2 dataset. STP represents the Static Token Pruning method, and MAG denotes our Multi-modality Alignment Guidance module.}
  \label{fig:results}
  \vspace{-8mm}
\end{figure}

To release this limitation, a few works have attempted to accelerate the VLT models. As a pioneer, Upop~\cite{2023Upop} suggests a unified parameter pruning strategy for compressing VLTs, allowing for simultaneous pruning of submodules across diverse modalities. 
Recently, considering the token number plays a dominant role in the total computation cost, several studies have put more effort into accelerating VLTs via pruning tokens. ELIP~\cite{guo2023elip} introduces a vision token pruning method to remove less influential tokens based on the supervision of language outputs. CrossGET~\cite{shi2023crossget} implements token pruning by selectively eliminating redundant tokens at each layer of the VLTs. Despite some progress achieved by these works, there still exist two unresolved issues. As depicted in Table~\ref{tab:comparison}, all these methods face challenges in exploring multi-modality alignment and different inputs to dynamically compress VLT, details are as follows.

\begin{table}[t]
    \centering
    \small
    \setlength{\tabcolsep}{0.6mm}{
    \begin{tabular}{@{}l|cccc@{}}
    \toprule
        Methods & \multicolumn{1}{c|}{\begin{tabular}[c]{@{}c@{}}Layer-wise \\ Dynamic\end{tabular}} & \multicolumn{1}{c|}{\begin{tabular}[c]{@{}c@{}}Instance-wise \\ Dynamic\end{tabular}} & \multicolumn{1}{c|}{\begin{tabular}[c]{@{}c@{}}Modality \\ Guidance\end{tabular}} & \multicolumn{1}{c}{\begin{tabular}[c]{@{}c@{}}Modality \\ Alignment\end{tabular}}  \\ \midrule
        Upop~\cite{2023Upop} & \ding{51} & \ding{55} & \ding{55} & \ding{55}   \\
        ELIP~\cite{guo2023elip} & \ding{55} & \ding{55} & \ding{51} & \ding{55}   \\
        CrossGET~\cite{shi2023crossget} & \ding{55} & \ding{55} & \ding{51}  & \ding{55}  \\ \midrule
        MADTP & \ding{51} & \ding{51} & \ding{51} & \ding{51}  \\
    \bottomrule
    \end{tabular}
    }
    \vspace{-3mm}
    \caption{Characteristics of existing compression methods for VLTs. The proposed MADTP first conducts visual-language modality alignment and then utilizes the aligned features to guide layer-wise and instance-wise dynamic token pruning.}
    \label{tab:comparison}
    \vspace{-6mm}
\end{table}

Firstly, existing popular VLT models~\cite{CLIP,kim2021vilt,li2022blip} usually consist of multiple modality-specific sub-modules for better capturing the representative knowledge for each modality, which often leads to imbalanced distributions of parameters and features between different modalities. Such imbalances have been extensively analyzed in studies~\cite{peng2022balanced, Fan_2023_CVPR}. In other words, different modality branches in VLT generally produce tokens with different representation capabilities for the same semantic concept. As a result, directly applying existing unimodal pruning methods~\cite{gordon2020compressing,zheng2022savit,patchslimming2021} to prune the VLT without considering each token's cross-modality semantic relevance, may falsely remove tokens that are less important in one modality but may be crucial in another. This will further worsen the representation capability imbalance between different modality branches in the compressed VLT. Thus, introducing cross-modality alignment can explicitly align the joint representation of different modalities for the same semantic concept, and increase the chances of eliminating less important tokens for all modalities, resulting in more effective compression of VLTs. 

Secondly, different input samples often require different levels of computation complexity~\cite{han2021dynamic,xia2021FDI} for inference. Hence, some research on unimodal dynamic token pruning~\cite{Chavan2022vitslimming,Meng2022adavit,Liu_Wu_Guo_2022,Yin_2022_CVPR} have emerged recently. These works offer flexibility in removing redundant tokens across different layers of the network by considering the complexity of input instances. However, one disadvantage is that these dynamic pruning works focus on single-modality compression, lacking the consideration of how to dynamically determine one token's importance across multi-modalities for different inputs. Another challenge is that, although promising, the exploration of dynamic token pruning for multimodal models is rarely studied. Thus, based on the aligned multi-modalities representations mentioned above, we further introduce dynamic token pruning modules at different layers of the Vision-Language Transformers, to achieve both input instance- and layer-wise VLT compression. 

In this work, we introduce a novel framework called Multimodal Alignment-Guided Dynamic Token Pruning (MADTP) to accelerate VLTs. 
The MADTP framework accepts image and text inputs, which are fed into a vision branch and a language branch to extract visual and language tokens, respectively. 
Then, the Multi-modality Alignment Guidance (MAG) module is designed to learn the semantic relevance between tokens from two modalities. Specifically, MAG utilizes learnable tokens to facilitate cross-modal feature alignment and guide the multimodal token pruning.
Furthermore, the Dynamic Token Pruning (DTP) module is presented within the Transformer blocks, enabling dynamic adjustment of the compression ratio for each layer based on the complexity of different input instances and the learned alignment guidance.
Fig.~\ref{fig:results} illustrates the substantial performance improvement achieved by our MADTP framework.


\noindent
Our main contributions can be summarized as follows:
\begin{itemize}[nosep,left=1em]
    \item We reveal the vital role of aligning multi-modalities for guiding VLT compression, and further propose a novel multimodal alignment-guided dynamic token pruning framework called MADTP, to effectively accelerate various Vision-Language Transformers.
    \item To relieve the unaligned modalities issue, we propose the Multi-modality Alignment Guidance (MAG) module, explicitly aligning the joint representations from different modalities and providing guidance during the multimodal token pruning process.
    \item To achieve adaptive VLT acceleration based on different inputs, we present the Dynamic Token Pruning (DTP) module, which dynamically adjusts the compression ratio for each layer of VLT models based on the complexity of input instance. 
    \item Extensive experiments across diverse datasets and models consistently verify that MADTP can achieve new state-of-the-art performance. Notably, MADTP achieves outstanding compression on the BLIP model in the NLVR2 dataset, reducing GFLOPs by 80\% while experiencing a performance decrease of less than 4\%.
\end{itemize}
\vspace{-1mm}
\section{Related Work}
\label{sec:Related Work}
\vspace{-1mm}
\subsection{Vision-Language Transformer}
\vspace{-1mm}
Vision-Language Transformer(VLT) models aim to make full use of information from different modalities and have been proven to be effective in various fields. CLIP~\cite{CLIP} and BLIP~\cite{li2022blip} are two representative VLT models. CLIP performs well on many downstream tasks by pretraining with images and texts matching. Further, BLIP uses a cross-attention layer to interact visual information with text information during the matching process of images and texts. Although VLT models show the powerful ability, they generally suffer high computation costs due to the need to process different modalities of information. Thus, it is necessary and of practical value to compress VLT models.

\vspace{-1mm}
\subsection{Multimodal Compression}
\vspace{-1mm}
The dominant techniques for model compression~\cite{han2015deep,cheng2017survey,sprechmann2015learning} encompass pruning~\cite{2015Structured,blalock2020state,Yang_Yin_Shen_Molchanov_Li_Kautz_2021,patchslimming2021,Yu_Xiang_2023}, quantization~\cite{gholami2022survey}, knowledge distillation~\cite{gou2021knowledge} and low-rank decomposition~\cite{yu2017compressing}, among others~\cite{iandola2016squeezenet,howard2017mobilenets}. However, these methods mainly focus on single-modality model compression, such as ViTs, while multimodal compression such as VLTs remain challenges. To this end, a few works have attempted to compress the VLT models recently.
As the pioneering work, DistillVLM~\cite{Fang2021DistillVLM} leverages knowledge distillation to transfer the knowledge from larger VLTs to smaller VLTs. 
Upop~\cite{2023Upop} adopts a layer-wise dynamic parameter pruning approach, which uniformly searches subnets and adaptively adjusts the pruning ratio of each layer.
ELIP~\cite{guo2023elip} presents a vision token pruning technique that eliminates less important tokens by leveraging language outputs as supervision.
CrossGET~\cite{shi2023crossget} introduces the cross tokens to facilitate multimodal token pruning.
However, all these methods overlook the significance of multi-modality alignment guidance for VLT compression, leading to a decrease in the performance of the compressed models.
Although some works~\cite{guo2023elip,shi2023crossget} attempt to utilize modality guidance to assist token pruning, this problem still exists. 
Our proposed MAG module explicitly aligns the feature representations of the two modalities using learnable tokens. It provides comprehensive guidance for subsequent dynamic token pruning process, enabling effective resolution of this challenge.

\vspace{-1mm}
\subsection{Token Merging and Pruning}
\vspace{-1mm}
Token merging and pruning~\cite{Bolya_Fu_Dai_Zhang_Feichtenhofer_Hoffman,Bian_Wang_Han_Wang_2023} are proven effective for model compression. ToMe~\cite{Bolya_Fu_Dai_Zhang_Feichtenhofer_Hoffman} designed a token merging strategy for ViTs, merging similar parts in each block. Further,~\cite{Bian_Wang_Han_Wang_2023} merges non-critical tokens into crucial tokens, which not only reduces the number of tokens but also retains more information. Most of these methods reduce a fixed number of tokens at each step. However, according to~\cite{Rao2021dynamicvit,Yin_Vahdat_Alvarez_Mallya_Kautz_Molchanov,Tang_Wang_Kong_Zhang_Li_Ding_Wang_Liang_Xu_2022}, the number of tokens retained by the current block should be related to its importance to the final task. 
DynamicViT~\cite{Rao2021dynamicvit} uses a prediction module to measure the importance of each patch embedding in the current input to decide whether to discard the patch. AdaViT~\cite{Yin_Vahdat_Alvarez_Mallya_Kautz_Molchanov} adaptively stop some tokens from participating in subsequent calculations. MuE~\cite{Tang_Wang_Kong_Zhang_Li_Ding_Wang_Liang_Xu_2022} design an early exiting strategy based on input similarity for ViT models. 
Unlike these works processing unimodal ViT models, we focus on reducing the computation cost of various VLT models, by designing a multimodal dynamic token pruning strategy based on the complexity of the input image and text pairs.

\begin{figure*}[t]
  \centering
  \includegraphics[width=0.96\linewidth]{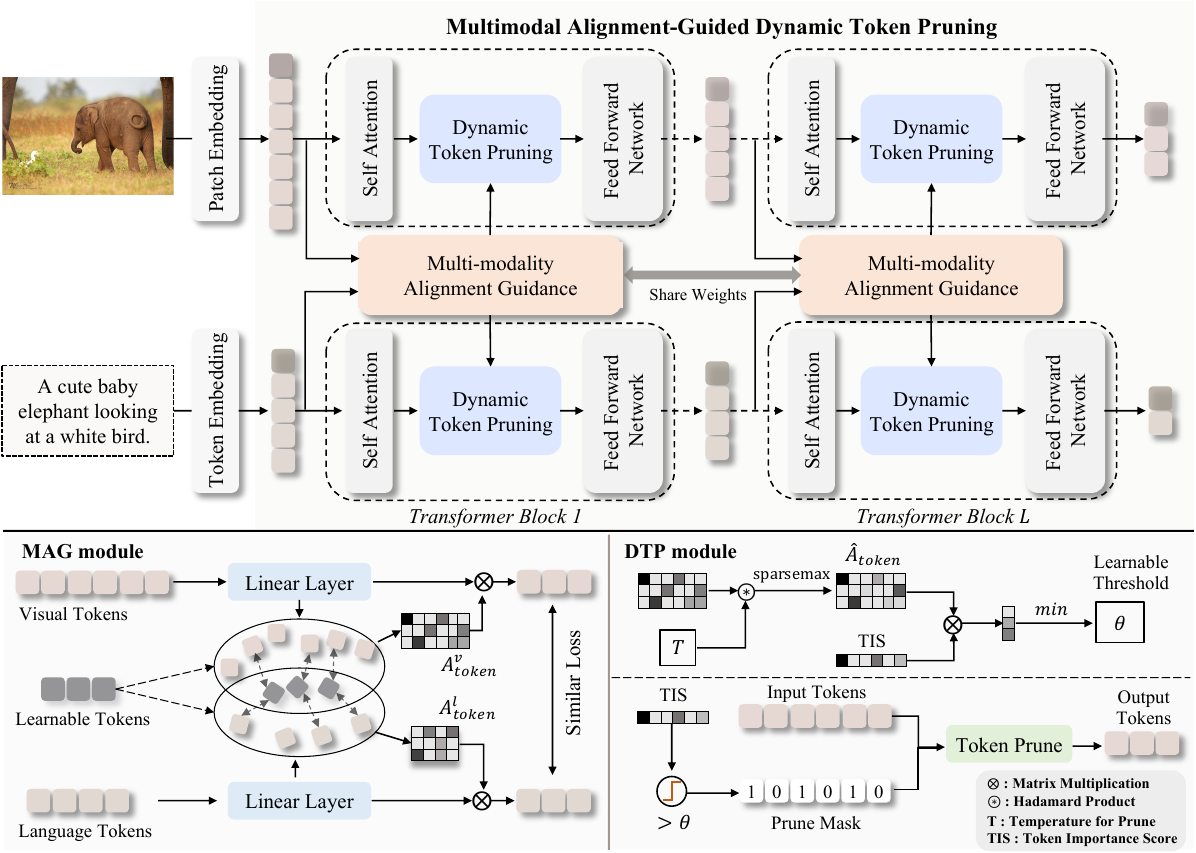}
  \vspace{-3mm}
  \caption{Overview of the proposed MADTP framework. It comprises two main components: the Multi-modality Alignment Guidance (MAG) module and the Dynamic Token Pruning (DTP) module. The MAG module is placed between the vision and language branches in VLTs, facilitating explicit alignment of representations across modalities and offering guidance for token pruning. Meanwhile, the DTP module is incorporated within each transformer block, allowing for dynamic token pruning based on the complexity of input instances.
  }
  \vspace{-6.5mm}
  \label{fig:framework}
\end{figure*}

\section{Methodology}
\label{sec:Methodology}
The MADTP architecture overview is depicted in Fig.~\ref{fig:framework}. In this following, we first give a brief introduction of the Vision-Language Transformers in Sec.~\ref{sec:VLT}. We then present our Multi-modality Alignment Guidance module and Dynamic Token Pruning module in Sec.~\ref{sec:MAG} and Sec.~\ref{sec:DTP}, respectively. Finally, we elaborate on the optimization function of the framework in Sec.~\ref{sec:OF}.

\subsection{Preliminaries}
\label{sec:VLT}
Vision-Language Transformers have emerged as the prominent architectures~\cite{CLIP,li2022blip,li2023blip} in multimodal learning, comprising two branches: the vision branch and the language branch.
The vision branch usually employs the ViT~\cite{vit2020image} as the visual encoder, while the language branch utilizes BERT~\cite{bert2019} as the language encoder, extracting visual and language tokens from their respective modalities. 
In detail, given an image and a text as inputs, the visual encoder performs patch embedding on the image to generate the visual tokens $V=\{V_1,V_2,...V_N\}$, where $N$ is the patch number, and the language encoder processes the words in the text using token embedding, converting them into language tokens $L=\{L_1,L_2,...,L_M\}$, where $M$ is the number of words. Furthermore, two learnable tokens, $V_{cls}$ and $L_{eos}$, are added to the visual tokens and language tokens, respectively.
These token embeddings provide comprehensive representations for the image and text inputs, which are then passed through transformer blocks for feature encoding. 
In VLTs, both the vision and language branches consist of $\text{L}$ layers of transformer blocks. Each block comprises a Multi-Head Self Attention (MHSA) layer and a Feed Forward Network (FFN) layer, enabling the model to capture contextual relationships within each modality.
In addition, some VLT models like BLIP~\cite{li2022blip}, incorporate several Cross Attention layers to capture inter-modal interactions and enhance information fusion between two modalities. 

\subsection{Multi-modality Alignment Guidance}
\label{sec:MAG}
As discussed in Sec.~\ref{sec:intro}, the unaligned modalities issue highlights the challenge of directly applying unimodal token pruning methods to VLTs. 
To alleviate this problem, the Multi-modality Alignment Guidance (MAG) module is designed to explicitly align the feature representations between two modalities, and provide sufficient guidance for the multimodal token pruning process. 
As shown in Fig.~\ref{fig:framework}, we insert the MAG module between the transformer blocks of two modal branches in the VLT architecture. 

Specifically, we first apply two linear layers to map the visual tokens $V$ and language tokens $L$ from each layer of VLTs into the same feature dimension. The linear layers and mapping process can be represented as follows:
\begin{equation}
    \begin{aligned}
    & V' = W_v  V + B_v, \\
    & L' = W_t  L + B_t,
    \end{aligned}
\end{equation}
where $V'$ and $L'$ are the mapped visual and language tokens, respectively. The $W_v$, $W_t$, $B_v$, and $B_t$ are layer-specific trainable weight matrices and biases.

Next, we utilize learnable tokens $E=\{E_1,E_2,...E_K\}$ as common feature space to establish associations between the visual and language modalities, where $K$ is the number of learnable tokens. In detail, we employ a scaled dot-product attention layer to calculate the correlation between the learnable tokens $E$ and the mapped visual tokens $V'$, resulting in token attention maps $A_{token}^v \in \mathbb{R}^{K \times N}$ and visual features $E^v$.
This process can be expressed as:
\begin{equation}
    \begin{aligned}
    A_{token}^v = softmax(\frac{E{V'}^T }{\sqrt[]{d_k}}) ,
    \end{aligned}
\end{equation}
\begin{equation}
    \begin{aligned}
    E^v = A_{token}^v * V' ,
    \end{aligned}
\end{equation}
where $d_k$ is a scaling factor. 
Similarly, we can also obtain the token attention maps $A_{token}^l \in \mathbb{R}^{K \times M}$ between the mapped language tokens and learnable tokens, and extract the language features $E^l$.

\begin{figure}[t]
  \centering
  \includegraphics[width=\linewidth]{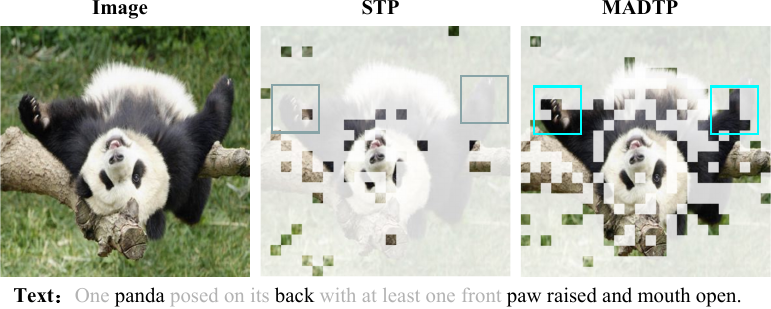}
  \vspace{-7mm}
  \caption{Visualization of token pruning results between STP and MADTP, providing strong evidence that our approach emphasizes modality correlation and effectively avoids pruning crucial tokens.}
  \label{fig:compaire_stp}
  \vspace{-5mm}
\end{figure}

Further, we calculate the similarity between these two features and incorporate it into the final loss constraint to assist the model during training. We believe that the visual and language features learned by the same learnable tokens should exhibit strong semantic relevance. Through the above operations, we explicitly align the representations between two modalities and obtain token attention maps representing the modality alignment achieved by the learnable tokens.
Afterward, these maps are fed into the Dynamic Token Pruning module to guide the token pruning process of the VLTs, ensuring that the pruned tokens are redundant in both modalities and enhancing the compression effectiveness of the multimodal model, which is exemplified in Fig.~\ref{fig:compaire_stp}. Note that the MAG modules share weights in the MADTP framework.

\subsection{Dynamic Token Pruning}
\label{sec:DTP}
Dynamic token pruning in single-modality compression has been proven to be more efficient than static token pruning, as it enables adaptive adjustment of the model's compression rate based on the complexity of the input instance.
Motivated by this, we have also designed a Dynamic Token Pruning (DTP) module in the MADTP framework.
As illustrated in Fig.~\ref{fig:framework}, we insert the DTP module between the Self Attention layer and the Feed Forward Network in each Transformer block, allowing it to dynamically reduce the number of input tokens at each layer of VLTs. 
Following a similar procedure as in the single-modality token pruning, we first calculate the importance score for each token. 
Then, a learnable threshold is employed to dynamically prune tokens at both the input instance-wise and layer-wise levels.

\noindent
\textbf{Token Importance Score.}
Apart from considering token importance based on the class attention map~\cite{Liu_Wu_Guo_2022,Yin_2022_CVPR}, as commonly done in traditional token pruning for ViTs, our approach extends to incorporate the importance of tokens within the same modality and the guidance of token alignment across different modalities. The Token Importance Score ($\text{TIS}$) is obtained by averaging three types of scores:
\begin{equation}
    \begin{aligned}
    \text{TIS} = (S_{\text{cls}} + S_{\text{self}} + S_{\text{token}}) / 3 ,
    \end{aligned}
    \label{eq:4}
\end{equation}
where $S_{\text{cls}}$ represents the class attention score as implemented by~\cite{Liu_Wu_Guo_2022}. 
$S_{\text{self}}$ and $S_{\text{token}}$ denote the self-attention score and token attention score, respectively.
Taking the visual modality as an example, we utilize the self-attention maps $A^v_{self}\in \mathbb{R}^{N \times N}$ from the MHSA layer and the token attention maps $A^v_{token}\in \mathbb{R}^{K \times N}$ obtained from the MAG module to calculate the attention scores $S^v_{\text{self}}$ and $S^v_{\text{token}}$ through the following steps:
\begin{equation}
    \begin{aligned}
    S_{\text{self}}^{v,k} = \frac{\max(A^{v,k}_{self})}{\sum_{k=1}^{N}\max(A^{v,k}_{self})} ,
    \end{aligned}
\end{equation}
\begin{equation}
    \begin{aligned}
    S_{\text{token}}^{v,k} = \frac{\max(A^{v,k}_{token})}{\sum_{k=1}^{N}\max(A^{v,k}_{token})} .
    \end{aligned}
\end{equation}
Here, $N$ refers to the total number of visual tokens. 
$\max(A^{v,k}_{self})$ and $\max(A^{v,k}_{token})$ represent the maximum value for the $k$-th token in the self-attention maps and token attention maps, respectively. 
To ensure the scores are within the range of $[0, 1]$, the attention scores ($S^v_{\text{self}}$ and $S^v_{\text{token}}$) are normalized by dividing them by the sum of their corresponding values.
Note that by incorporating these three attention scores, our $\text{TIS}$ can effectively avoid discarding crucial tokens by considering their relevance to the task, as well as their importance within and across modalities.

\noindent
\textbf{Learnable Threshold.}
To achieve instance-wise adaptive token pruning while minimizing operational costs, we propose the use of learnable thresholds for dynamic token pruning within MADTP. 
Specifically, we utilize the token attention maps $A_{token}$ learned from the MAG module to compute these thresholds. 
Firstly, we multiply $A_{token}$ by a temperature parameter $T$ and apply $\rm{sparsemax}$ function~\cite{martins2016softmax} to obtain sparse token attention maps, denoted as $\hat{A}_{token}$, 
\begin{equation}
    \begin{aligned}
    \hat{A}_{token} = {\rm sparsemax}(T * A_{token}) .
    \end{aligned}
\end{equation}

The role of the $\rm{sparsemax}$ function is to produce sparse distributions by minimizing the squared Euclidean distance between the output distribution and the input values.
\begin{equation}
    \begin{aligned}
    {\rm sparsemax}(\boldsymbol{z}):=\mathop{\arg\min}_{\boldsymbol{p} \in \Delta^{K-1}} \ \ \|\boldsymbol{p} - \boldsymbol{z} \|^{2} ,
    \end{aligned}
\end{equation}
where $\Delta^{K-1} := \{\boldsymbol{p} \in \mathbb{R}^{K} | \boldsymbol{1}^{T}\boldsymbol{p} = 1, \boldsymbol{p} \geq 0\}$.
Next, we perform matrix multiplication between $\hat{A}_{token}$ and $\text{TIS}$ to obtain $K$ thresholds, and take the minimum value among these thresholds as the final threshold $\theta$, used for the following token pruning procedure for this DTP module. 
\begin{equation}
    \begin{aligned}
    \theta = \min(\hat{A}_{token} \otimes \text{TIS}) .
    \end{aligned}
\end{equation}

\noindent
\textbf{Token Pruning.}
Based on the token importance scores and learnable threshold mentioned above, we can proceed with the designed token pruning scheme to reduce the number of input tokens. 
Firstly, we compare the $\text{TIS}$ score of each token with the threshold $\theta$ to obtain the prune mask $M_p$, which can be formulated in Equation~\ref{eq}:
\begin{equation}
    \label{eq}
    \begin{aligned}
    M_p(x_i) =\left\{
        \begin{aligned}
            1 & , && if \ \ \text{TIS}(x_i) > \theta , \\
            0 & , && otherwise .
        \end{aligned}
    \right. 
    \end{aligned}
\end{equation}
Where $x_i$ represents the $i$-th input tokens. 
Then we keep the tokens with scores greater than the threshold and eliminate the other tokens according to the pruning mask.
However, directly discarding tokens may result in information loss. To address this, we adopt a similar approach as EVit~\cite{liang2022evit}, weighting the pruned tokens based on their $\text{TIS}$ to generate a new token, which is then added to the retained tokens.

\subsection{Objective Function}
\label{sec:OF}
Due to VLTs having different loss functions for various multimodal tasks, 
we represent the specific task loss function as $L_{task}$ during training. 
Additionally, as explained in Section~\ref{sec:MAG}, we incorporate a similar loss denoted as $L_{sim}$ to capture the alignment relationship between the visual features $E^v$ and language features $E^l$ obtained from the MAG modules for optimizing the model pruning process. Consequently, the overall loss function $L$ of the proposed MADTP framework can be expressed as:
\begin{equation}
    L = L_{task} + \alpha L_{sim} ,
\end{equation}
where $\alpha$ denotes the balance coefficient. The computation for $L_{sim}$ is defined as follows:
\begin{equation}
    L_{sim} = \frac{1}{K}\sum_{i=1}^{K}(1 - cos(E^v_i, E^l_i)) .
\end{equation}
Where $K$ is the number of visual and language features. 
\section{Experiments}
\label{sec:Experiments}
\vspace{-1mm}
\subsection{Experimental Setup}
\vspace{-1mm}
\noindent
\textbf{Dataset and evaluation metrics.} To evaluate our method comprehensively, four multimodal datasets are used, including NLVR2~\cite{nlvr2}, COCO~\cite{Imagecaption_2014}, Flickr30k~\cite{flick30k} and VQA v2.0~\cite{vqa}. NLVR2~\cite{nlvr2} contains 107,292 pairs of images and text descriptions. COCO~\cite{Imagecaption_2014} comprises around 330,000 images, each accompanied by five text descriptions.
Flickr30k~\cite{flick30k} is mainly used for image and text retrieval tasks, and consists of 31,783 images, and each image has a descriptive title. VQA v2.0~\cite{vqa} is a human-annotated, open-ended question-and-answer dataset about images.
Performance evaluation metrics are task-specific, while model complexity is measured in GFLOPs (Giga-Floating-Operations per image-text pair). Please refer to the Appendix A for more details.

\noindent
\textbf{Implementation details.}
We use the MADTP framework to compress the CLIP~\cite{CLIP} and BLIP~\cite{li2022blip} models, which are initialized with pretrained weights from the official implementation of ~\cite{2023Upop}. 
During the compressing process, we utilize $8$ A100 GPUs with a batch size of $32$, and the hyper-parameter $\alpha$ in the loss function is set to $0.1$. The temperature $T$ in the DTP module is dynamically adjusted at each epoch, based on the GFLOPs of the pruned model.
Due to space limitations and the variability of training configurations across different models, more detailed experiment settings can be found in Appendix B. 

\vspace{-1mm}
\subsection{Experiments on the Visual Reasoning Task}
\begin{figure*}[t]
  \centering
  \includegraphics[width=0.98\linewidth]{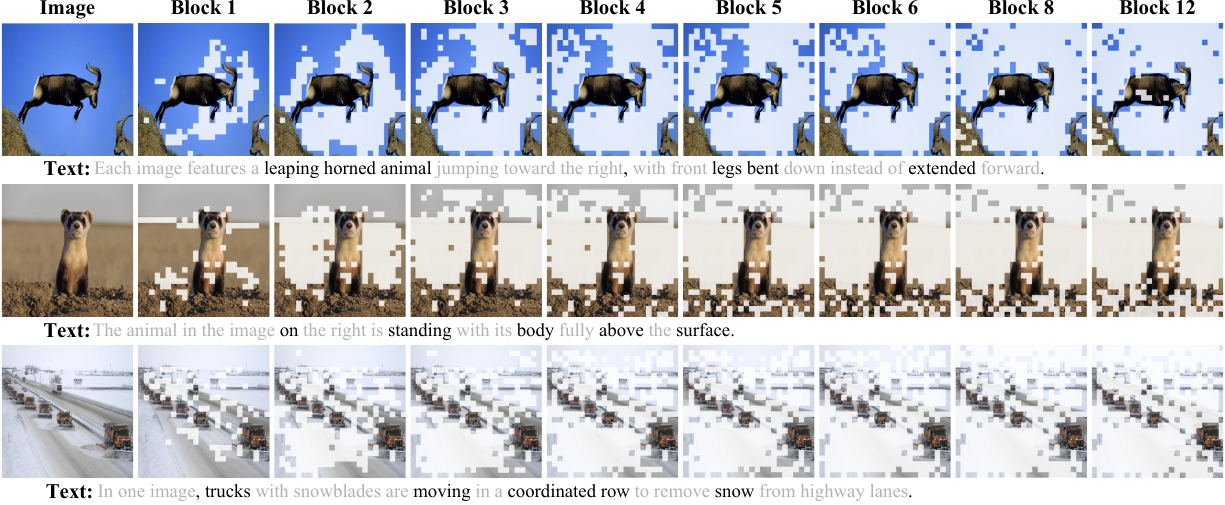}
  \vspace{-3mm}
  \caption{
  Visualization of our MADTP’s compressed BLIP results on NLVR2 dataset at each transformer block. The white mask in the image represents the pruned visual tokens, while the gray words in the text indicate the discarded language tokens. Our method effectively learns semantic relevance between modalities and effectively prunes tokens that are unimportant in both modalities.
  }
  \label{fig:Visualization}
  \vspace{-3.2mm}
\end{figure*}

\vspace{-1mm}
In this section, we conduct experiments utilizing our MADTP framework to compress the BLIP model on the NLVR2 dataset. 
In Table \ref{tab:blip_nlvr2}, we compare our approach with the state-of-the-art method~\cite{2023Upop} to demonstrate its effectiveness.
Additionally, we perform ablation studies to analyze the impact of different components and hyperparameters of the MADTP framework, presenting the results in Table~\ref{tab:component_MADTP} and Table~\ref{tab:hyper_MADTP}, respectively.
Moreover, we visualize the token pruning results for the compressed model in Fig.~\ref{fig:Visualization}.

\begin{table}[t]
\small
\centering
\setlength{\tabcolsep}{1.5mm}{
\begin{tabular}{c|c|cc|l}
\toprule
 &  &  &  & \\
\multirow{-2}{*}{Approach} & \multirow{-2}{*}{\makecell{Reduce \\ Ratio}} & \multirow{-2}{*}{Dev Acc}& \multirow{-2}{*}{Test Acc} & \multirow{-2}{*}{GFLOPs} \\
\midrule
Uncompressed& /& 82.48& 83.08& 132.54\\ \midrule
            & 0.3 & 79.50& 80.01& 94.08
            \\
\multirow{-2}{*}{STP} & 0.5& 78.08& 77.61& 68.31
\\ \midrule
            & 0.3 & 80.33& 81.13& 89.36
            \\
            & 0.5& 76.89& 77.61& 65.29
            \\
            & 0.6& 72.85& 73.55& 50.35
            \\
            & 0.7& 68.71& 68.76& 39.93
            \\
\multirow{-5}{*}{UPop~\cite{2023Upop}}& 0.8  & 57.17& 57.79& 19.08
\\ \midrule
            & 0.3& \textbf{82.50}& \textbf{83.20}&  92.60{\scriptsize\color[HTML]{FE0000}$\downarrow$30\%} \\  
            & 0.5& \textbf{81.97}& \textbf{82.85}& 66.16{\scriptsize\color[HTML]{FE0000}$\downarrow$50\%}  \\
            & 0.6& \textbf{81.92}& \textbf{82.42}& 52.92{\scriptsize\color[HTML]{FE0000}$\downarrow$60\%}  \\ 
            & 0.7& \textbf{80.67}& \textbf{81.23}& 39.69{\scriptsize\color[HTML]{FE0000}$\downarrow$70\%}  \\ 
\multirow{-5}{*}{\textbf{\makecell{MADTP \\ (Ours)}}} & 0.8& \textbf{78.28}& \textbf{79.22}& 26.46{\scriptsize\color[HTML]{FE0000}$\downarrow$80\%} \\ 
\bottomrule
\end{tabular}
}
\vspace{-2mm}
\caption{Comparison of compression results for BLIP model on the NLVR2 dataset. Bold indicates the best results. Reduce Ratio indicates the desired compression ratio of GFLOPs.}
\vspace{-3mm}
\label{tab:blip_nlvr2}
\end{table}

\begin{table}[ht]
\small
\centering
\setlength{\tabcolsep}{1.5mm}{
\begin{tabular}{cl|cc|c}
\toprule
\multicolumn{2}{c|}{Components of MADTP} & Dev Acc  & Test Acc & GFLOPS \\ \midrule
& only w/ $S_{\text{self}}$    & 81.49   & 82.13  & 70.46  \\ 
& only w/ $S_{\text{token}}$   & 80.68   & 81.00  & 66.74  \\ 
\multirow{-3}{*}{$\text{TIS}$} & only w/ $S_{\text{cls}}$    & 81.62   & 82.25  & 69.67  \\ \midrule
& w/o MAG & 79.65 & 80.96 & 68.91 \\
\multirow{-2}{*}{Module} & w/o DTP & 80.83 & 81.44 & 68.70\\ \midrule
\multicolumn{2}{c|}{\textbf{MADTP (Ours)}} & \textbf{81.97}  & \textbf{82.85}  & \textbf{66.16}  \\
\bottomrule
\end{tabular}
}
\vspace{-2mm}
\caption{
Ablation study of different components in MADTP framework for compressing BLIP on NLVR2 at $0.5$ reduce ratio.
}
\vspace{-3mm}
\label{tab:component_MADTP}
\end{table}

\begin{table}[ht]
\small
\centering
\setlength{\tabcolsep}{1.8mm}{
\begin{tabular}{cl|cc|c}
\toprule
\multicolumn{2}{c|}{Hyperparameters} & Dev Acc & Test Acc & GFLOPS         \\ \midrule
 & 50  & 81.44  & 82.03   & 67.70 \\  
 & \textbf{100}   & \textbf{81.97}  & \textbf{82.85} & \textbf{66.16}\\  
 & 150   & 81.49  & 82.19 & 66.79 \\  
\multirow{-4}{*}{$K$}  & 200   & 81.74  & 81.96             & 66.99 \\ \midrule
& 256   & 81.79   & 82.28    & 66.94\\  
& 512  & 81.79  & 82.46    & 68.63 \\  
& \textbf{768}   & \textbf{81.97}   & \textbf{82.85}          & \textbf{66.16} \\  
\multirow{-4}{*}{$d_k$} & 1024  & 81.60  & 81.95             & 66.55 \\  \midrule
& mean-keep  & 81.34  & 81.70  & 67.10 \\
\multirow{-2}{*}{\makecell{Operation}} & \textbf{max-keep}   & \textbf{81.97}  & \textbf{82.85}  & \textbf{66.16}  \\ 
\bottomrule
\end{tabular}
}
\vspace{-3mm}
\caption{Hyperparameters for compressing BLIP on NLVR2 at 0.5 reduce ratio. $K$ and $d_k$ donets the number and the channel dimension of learnable tokens. The "mean-keep" and "max-keep" operations are utilized for parallel training within each mini-batch.}
\vspace{-6.5mm}
\label{tab:hyper_MADTP}
\end{table}

\begin{table*}[hbt!]
\small
\centering
\setlength{\tabcolsep}{3mm}{
\begin{tabular}{c|c|c|ccc|ccc|l}
\toprule
&  &  &  \multicolumn{3}{c|}{Image$\to$Text}& \multicolumn{3}{c|}{Text$\to$Image}&  \\ 
\multirow{-2}{*}{Dataset}  & \multirow{-2}{*}{Approach} & \multirow{-2}{*}{\makecell{Reduce \\ Ratio}} & \multicolumn{1}{c}{R@1}   & \multicolumn{1}{c}{R@5}   & \multicolumn{1}{c|}{R@10} & \multicolumn{1}{c}{R@1}   & \multicolumn{1}{c}{R@5}   & \multicolumn{1}{c|}{R@10}& \multirow{-2}{*}{GFLOPS}           \\ \midrule
                             & Uncompressed & /& \multicolumn{1}{c}{96.8}& \multicolumn{1}{c}{100.0}& \multicolumn{1}{c|}{100.0}& \multicolumn{1}{c}{86.6} & \multicolumn{1}{c}{97.8}& \multicolumn{1}{c|}{99.1}&395.7 \\ \cline{2-10}  
                             & &0.5& \multicolumn{1}{c}{93.2}& \multicolumn{1}{c}{99.4}& \multicolumn{1}{c|}{99.8}& \multicolumn{1}{c}{80.5}& \multicolumn{1}{c}{95.4}& \multicolumn{1}{c|}{97.6}& 201.1
                             \\ 
                             & \multirow{-2}{*}{UPop~\cite{2023Upop}} &0.75& \multicolumn{1}{c}{82.9}& \multicolumn{1}{c}{95.7}& \multicolumn{1}{c|}{97.8}& \multicolumn{1}{c}{67.3}&\multicolumn{1}{c}{89.5}& \multicolumn{1}{c|}{93.5}& 102.6
                             \\   \cline{2-10} 
                             & & 0.5& \multicolumn{1}{c}{\textbf{93.9}}& \multicolumn{1}{c}{\textbf{99.5}}& \multicolumn{1}{c|}{\textbf{99.8}}& \multicolumn{1}{c}{\textbf{83.3}} & \multicolumn{1}{c}{\textbf{97.0}} & \multicolumn{1}{c|}{\textbf{98.5}}& 178.8{\scriptsize\color[HTML]{FE0000}$\downarrow$55\%} \\              
\multirow{-5}{*}{\makecell{ Flickr30K \\ (1K test set)}}& \multirow{-2}{*}{\textbf{MADTP (Ours)}} & 0.75& \multicolumn{1}{c}{\textbf{88.4}}  & \multicolumn{1}{c}{\textbf{97.3}}  & \multicolumn{1}{c|}{\textbf{99.0}}  & \multicolumn{1}{c}{\textbf{76.9}} & \multicolumn{1}{c}{\textbf{94.2}}  & \multicolumn{1}{c|}{\textbf{97.0}}& 99.5{\scriptsize\color[HTML]{FE0000} $\downarrow$75\%}  \\  \midrule
                             & Uncompressed& / & \multicolumn{1}{c}{71.5}& \multicolumn{1}{c}{90.8}& \multicolumn{1}{c|}{95.4} & \multicolumn{1}{c}{56.8}        & \multicolumn{1}{c}{80.7}& \multicolumn{1}{c|}{87.6}& 395.7     \\  \cline{2-10} 
                             & &0.5& \multicolumn{1}{c}{70.8}& \multicolumn{1}{c}{90.8}& \multicolumn{1}{c|}{95.2}& \multicolumn{1}{c}{53.1}& \multicolumn{1}{c}{79.9}  & \multicolumn{1}{c|}{87.3}& 196.3
                             \\   
                             & \multirow{-2}{*}{UPop~\cite{2023Upop}} & 0.75& \multicolumn{1}{c}{56.1}& \multicolumn{1}{c}{82.4}& \multicolumn{1}{c|}{90.2}& \multicolumn{1}{c}{41.1}& \multicolumn{1}{c}{71.0}& \multicolumn{1}{c|}{81.4}& 105.9
                             \\ \cline{2-10} 
                             & &0.5& \multicolumn{1}{c}{\textbf{72.7}} & \multicolumn{1}{c}{\textbf{91.8}} & \multicolumn{1}{c|}{\textbf{96.1}}& \multicolumn{1}{c}{\textbf{55.0}} & \multicolumn{1}{c}{\textbf{79.9}}& \multicolumn{1}{c|}{\textbf{87.5}}& 190.2{\scriptsize\color[HTML]{FE0000}$\downarrow$52\%} \\  
\multirow{-5}{*}{\makecell{ COCO \\ (5K test set)}} & \multirow{-2}{*}{\textbf{MADTP (Ours)}} & 0.75& \multicolumn{1}{c}{\textbf{66.2}} & \multicolumn{1}{c}{\textbf{88.4}} & \multicolumn{1}{c|}{\textbf{93.7}} & \multicolumn{1}{c}{\textbf{49.9}} & \multicolumn{1}{c}{\textbf{76.3}} & \multicolumn{1}{c|}{\textbf{85.1}} & 92.4{\scriptsize\color[HTML]{FE0000}$\downarrow$77\%}  \\ \bottomrule
\end{tabular}
}
\vspace{-2mm}
\caption{Compress CLIP on the Flickr30K and COCO datasets of the Image-Text Retrieval task. The R@1, R@5, and R@10 are the higher the better. The best results are in bold.}
\vspace{-3mm}
\label{tab:clip_flick30k_coco}
\end{table*}

\begin{table*}[hbt!]
\small
\centering
\setlength{\tabcolsep}{3mm}{
\begin{tabular}{c|c|c|ccc|ccc|l}
\toprule
&  &  & \multicolumn{3}{c|}{Image$\to$Text}& \multicolumn{3}{c|}{Text$\to$Image}&  \\ 
\multirow{-2}{*}{Dataset}  & \multirow{-2}{*}{Approach} & \multirow{-2}{*}{\makecell{Reduce \\ Ratio}} & \multicolumn{1}{c}{R@1}   & \multicolumn{1}{c}{R@5}   & \multicolumn{1}{c|}{R@10} & \multicolumn{1}{c}{R@1}   & \multicolumn{1}{c}{R@5}   & \multicolumn{1}{c|}{R@10}& \multirow{-2}{*}{GFLOPS}           \\ \midrule
                             & Uncompressed& /& \multicolumn{1}{c}{96.8}& \multicolumn{1}{c}{99.9}& \multicolumn{1}{c|}{100.0}& \multicolumn{1}{c}{86.9}& \multicolumn{1}{c}{97.3}& \multicolumn{1}{c|}{98.7}& 153.2 \\ \cline{2-10}   
                             & &0.5& \multicolumn{1}{c}{94.0}& \multicolumn{1}{c}{99.5}  & \multicolumn{1}{c|}{99.7}& \multicolumn{1}{c}{82.0}& \multicolumn{1}{c}{95.8}& \multicolumn{1}{c|}{97.6}& 91.0
                             \\ 
                             & \multirow{-2}{*}{UPop~\cite{2023Upop}}&0.75& \multicolumn{1}{c}{85.8}& \multicolumn{1}{c}{97.4}& \multicolumn{1}{c|}{98.4}& \multicolumn{1}{c}{71.3}& \multicolumn{1}{c}{91.0}& \multicolumn{1}{c|}{94.9}&51.0
                             \\  \cline{2-10} 
                             & & 0.5& \multicolumn{1}{c}{\textbf{95.1}}  & \multicolumn{1}{c}{\textbf{99.5}}& \multicolumn{1}{c|}{\textbf{99.7}}  & \multicolumn{1}{c}{\textbf{82.3}} & \multicolumn{1}{c}{\textbf{96.2}} & \multicolumn{1}{c|}{\textbf{98.0}} & 74.5{\scriptsize\color[HTML]{FE0000}$\downarrow$51\%}  \\  
\multirow{-5}{*}{\makecell{ Flickr30K \\ (1K test set)}} & \multirow{-2}{*}{\textbf{MADTP (Ours)}} & 0.75& \multicolumn{1}{c}{\textbf{91.8}}  & \multicolumn{1}{c}{\textbf{98.5}}  & \multicolumn{1}{c|}{\textbf{99.6}}  & \multicolumn{1}{c}{\textbf{77.1}} & \multicolumn{1}{c}{\textbf{93.2}} & \multicolumn{1}{c|}{\textbf{96.1}} & 58.7{\scriptsize\color[HTML]{FE0000}$\downarrow$62\%}\\ \midrule 
                             & Uncompressed& /& \multicolumn{1}{c}{81.9}& \multicolumn{1}{c}{95.4}& \multicolumn{1}{c|}{97.8}& \multicolumn{1}{c}{64.3}& \multicolumn{1}{c}{85.7}& \multicolumn{1}{c|}{91.5}& 153.2\\ \cline{2-10}  
                             & &0.5& \multicolumn{1}{c}{77.4}& \multicolumn{1}{c}{93.4}& \multicolumn{1}{c|}{97.0}& \multicolumn{1}{c}{59.8}& \multicolumn{1}{c}{83.1}& \multicolumn{1}{c|}{89.8}& 88.3
                             \\ 
                             & \multirow{-2}{*}{UPop~\cite{2023Upop}} &0.75& \multicolumn{1}{c}{62.9}& \multicolumn{1}{c}{86.2}& \multicolumn{1}{c|}{92.3}& \multicolumn{1}{c}{47.4}& \multicolumn{1}{c}{74.8}& \multicolumn{1}{c|}{83.9}& 50.2
                             \\  \cline{2-10} 
                             & & 0.5& \multicolumn{1}{c}{\textbf{79.1}} & \multicolumn{1}{c}{\textbf{94.2}} & \multicolumn{1}{c|}{\textbf{97.2}} & \multicolumn{1}{c}{\textbf{60.3}} & \multicolumn{1}{c}{\textbf{83.6}} & \multicolumn{1}{c|}{\textbf{89.9}}  & 87.4{\scriptsize\color[HTML]{FE0000} $\downarrow$43\%}  \\                     
\multirow{-5}{*}{\makecell{ COCO \\ (5K test set)}} & \multirow{-2}{*}{\textbf{MADTP (Ours)}} & 0.75 & \multicolumn{1}{c}{\textbf{71.2}} & \multicolumn{1}{c}{\textbf{90.0}} & \multicolumn{1}{c|}{\textbf{94.0}} & \multicolumn{1}{c}{\textbf{53.4}} & \multicolumn{1}{c}{\textbf{78.4}} & \multicolumn{1}{c|}{\textbf{86.2}} & 50.2{\scriptsize\color[HTML]{FE0000}$\downarrow$67\%}           \\ \bottomrule
\end{tabular}
}
\vspace{-2mm}
\caption{Compress BLIP on the Flickr30K and COCO datasets of the Image-Text Retrieval task. The R@1, R@5, and R@10 are the higher the better. The best results are in bold.}
\vspace{-6mm}
\label{tab:blip_flick30k_coco}
\end{table*}

\noindent
\textbf{Comparison to State-of-the-art Approaches.}
We report the performance of the MADTP framework for compressing the BLIP model at reduce ratios of 0.3, 0.5, 0.6, 0.7, and 0.8. The reduce ratio represents the proportion of the model's GFLOPs targeted for compression.
In order to assess the efficiency of our dynamic compression approach, we implement a baseline approach called Static Token Pruning (STP) which prunes a fixed number $k$ of redundant tokens at each layer of the VLTs  based on their importance scores computed in equation \ref{eq:4}.
In Table \ref{tab:blip_nlvr2}, under a reduce ratio of 0.3, MADTP achieved a 2.17\% increase in accuracy on the dev set and a 2.07\% increase on the test set compared to Upop~\cite{2023Upop}. Notably, at a reduce ratio of 0.5, these improvements extended to 5.08\% and 5.24\%, respectively.
Even at higher reduce ratios of 0.6, 0.7, and 0.8, MADTP demonstrated its ability to further compress the model while maintaining performance within an acceptable range. 
Remarkably, at a reduce ratio of 0.8, our method only experienced a 3.86\% drop on the test set compared to the uncompressed model.
These results highlight the effectiveness and superiority of our MADTP in achieving substantial model compression while preserving task performance across different reduce ratios.

\noindent
\textbf{Effect of Components.}
Table \ref{tab:component_MADTP} illustrates the contributions of different components in the proposed MADTP framework. We evaluate the impact of Token Importance Scores ($\text{TIS}$) and observe that combining scores from three sources yields the best results for token pruning. Additionally, we assess the individual effects of the two modules introduced in the MADTP framework. 
The MAG module improves performance by 2.32\% on the dev set and 1.89\% on the test set. Similarly, the DTP module leads to performance improvements of 1.14\% and 1.41\% on the respective sets. These experiments confirm the effectiveness of our proposed module within the MADTP framework.

\noindent
\textbf{Effect of Hyperparameters.} 
To illustrate the influence of various hyperparameters in the proposed MADTP framework,
we compare the performance of the pruned model under different hyperparameter settings. 
Table \ref{tab:hyper_MADTP} showcases how the compression results are influenced by the number and channel dimensions of learnable tokens in the MAG module.
The best performance is achieved when $K$ is set to $100$ and $d_k$ is set to $768$.
Additionally, we discuss the pruning strategy used in the dynamic token pruning process. 
The results indicate that the "max-keep" operation yields the best results, which determine the number of tokens to prune for a mini-batch based on the instance with the highest inference complexity.

\begin{table*}[hbt!]
\small
\centering
\setlength{\tabcolsep}{3mm}{
\begin{tabular}{c|c|ccl|ccl}
\toprule
\multirow{2}{*}{Approach} & \multirow{2}{*}{\makecell{Reduce \\ Ratio}} & \multicolumn{3}{c|}{Image Caption}                                                & \multicolumn{3}{c}{Visual Question Answering}                                          \\ 
                                   &                                        & \multicolumn{1}{c}{CIDEr} & \multicolumn{1}{c}{SPICE} & GFLOPs & \multicolumn{1}{c}{Test-dev} & \multicolumn{1}{c}{Test-std} & GFLOPs \\ \midrule
Uncompressed                       & /                                      & \multicolumn{1}{c}{133.3}          & \multicolumn{1}{c}{23.8}           & 65.7           & \multicolumn{1}{c}{77.4}              & \multicolumn{1}{c}{77.5}              & 186.1          \\ \midrule
 & 0.5                                   & \multicolumn{1}{c}{128.9}          & \multicolumn{1}{c}{23.3}           & 39.8
 & \multicolumn{1}{c}{76.3}     & \multicolumn{1}{c}{76.3}     & 109.4
\\ 
\multirow{-2}{*}{UPop~\cite{2023Upop}} & 0.75                                   & \multicolumn{1}{c}{117.4}          & \multicolumn{1}{c}{21.7}           & 22.2
& \multicolumn{1}{c}{74.5}              & \multicolumn{1}{c}{74.6}              & 62.3
\\ \midrule
 & 0.5                                   & \multicolumn{1}{c}{\textbf{131.0}} & \multicolumn{1}{c}{\textbf{23.5}}  & 39.7{\scriptsize\color[HTML]{FE0000} $\downarrow$39\%} & \multicolumn{1}{c}{\textbf{76.8}}             & \multicolumn{1}{c}{\textbf{76.8}}             & 79.4{\scriptsize\color[HTML]{FE0000} $\downarrow$57\%}  \\
\multirow{-2}{*}{\textbf{MADTP (Ours)}}  & 0.75                                  & \multicolumn{1}{c}{\textbf{120.1}} & \multicolumn{1}{c}{\textbf{22.0}}  & 22.1{\scriptsize\color[HTML]{FE0000} $\downarrow$66\%}          & \multicolumn{1}{c}{\textbf{76.3}}                 & \multicolumn{1}{c}{\textbf{76.2}}                 & 61.6{\scriptsize\color[HTML]{FE0000} $\downarrow$67\%}              \\ \bottomrule
\end{tabular}
}
\vspace{-3mm}
\caption{Compress BLIP on the Image Caption task and the Visual Question Answering task. The CIDEr, SPICE, test-dev, and test-std are the higher the better. The best results are in bold.}
\vspace{-6.5mm}
\label{tab:blip_imagecaption}
\end{table*}

\vspace{-1mm}
\subsection{Experiments on the Retrieval Task}
\vspace{-1mm}
We compress the CLIP~\cite{CLIP} and BLIP~\cite{li2022blip} models on the Flickr30K and COCO datasets with reduce ratios of 0.5 and 0.75, respectively. Tables \ref{tab:clip_flick30k_coco} and \ref{tab:blip_flick30k_coco} demonstrate the superior performance of our MADTP framework in image-text retrieval tasks across different model architectures.
It can be observed that when compressing the CLIP model on COCO dataset using our MADTP, there is a significant improvement in various metrics compared to the Upop~\cite{2023Upop}. Particularly, for high reduce ratio such as 0.75, we achieved improvements of up to 10\% in certain metrics (e.g., image-to-text recall@1 increased from 56.1\% to 66.2\%), and our GFLOPS metric is lower. 
Similarly, our MADTP compression experiments on the BLIP model also achieve impressive results compared to the Upop~\cite{2023Upop} method.

\vspace{-1mm}
\subsection{Experiments on the Image Caption Task}
\vspace{-1mm}
To assess the generalization capability of our proposed MADTP, we conducted additional experiments on the Image Caption task. Specifically, we compressed the BLIP model using reduce ratios of 0.5 and 0.75 on the COCO caption dataset. The results in Table~\ref{tab:blip_imagecaption} demonstrate the superior performance of our MADTP in the Image Caption task. Specifically, our MADTP method surpasses Upop~\cite{2023Upop} in terms of the CIDEr metric, achieving a 2.1\% improvement at a reduce ratio of 0.5 and a 2.7\% improvement at a reduce ratio of 0.75. These results emphasize the potential of MADTP in finding a balance between the computational cost of Vision-Language Transformers (VLTs) and maintaining high-quality image captioning capabilities.

\vspace{-1mm}
\subsection{Experiments on the Visual QA Task}
\vspace{-1mm}
In order to further validate the effectiveness of our MADTP method, we conducted compression experiments on the BLIP model using the VQA v2.0 dataset with reduce ratios of 0.5 and 0.75. The results, as depicted in Table \ref{tab:blip_imagecaption}, provide clear evidence that MADTP outperforms Upop~\cite{2023Upop} in terms of compression performance on the Visual QA task, particularly at higher reduce ratios. 
It is worth noting that our MADTP method achieves a remarkable 57\% reduction in the GFLOPs of the BLIP model while maintaining a performance degradation of less than 1\%.
These experimental findings serve as strong validation for the capability of our MADTP method to effectively accelerate VLTs while preserving model performance.

\vspace{-1mm}
\subsection{Discussion}
\vspace{-1mm}
Our MADTP can significantly reduce the computational costs of VLTs through token pruning, but does not reduce the models' parameters. To this end, we further verify the orthogonality of MADTP with parameter pruning methods, and the experimental results are provided in Appendix C. 
Our future work involves integrating a parameter pruning scheme into the proposed MADTP for comprehensive VLT model compression.

\vspace{-2mm}
\section{Conclusion}
\vspace{-1mm}
\label{sec:Conclusion}
We present the Multi-modality Alignment-Guided Dynamic Token Pruning (MADTP) framework to tackle the heavy computation costs of VLTs.
Our MADTP integrates the MAG module, which aligns features across modalities and guides the token pruning process to eliminate less important tokens in both modalities.
Additionally, the DTP module is introduced to dynamically adjust the token compression ratio based on complexity of input instance. 
Through extensive experiments, we show that MADTP is a promising approach for accelerating VLTs by reducing computational costs without sacrificing performance.
\vspace{-2mm}
\section{Acknowledgments}
This work is supported by National Natural Science Foundation of China (No. 62071127, and 62101137), National Key Research and Development Program of China (No. 2022ZD0160100), Shanghai Natural Science Foundation (No. 23ZR1402900).
\vspace{-2mm}
{
    \small
    \bibliographystyle{ieeenat_fullname}
    \bibliography{MADTP}
}

\clearpage
\renewcommand\thesection{\Alph{section}}
\setcounter{section}{0}
\setcounter{figure}{0}
\maketitlesupplementary
\renewcommand{\thefootnote}{\daggerfootnote{footnote}}

\section{Dataset and evalution metrics}
\label{sec:dataset}
We have conducted extensive experiments to evaluate our MADTP framework, utilizing four diverse multimodal datasets, namely NLVR2~\cite{nlvr2}, COCO~\cite{Imagecaption_2014}, Flickr30k~\cite{flick30k}, and VQA v2.0~\cite{vqa}. These datasets encompass a wide range of tasks and challenges, allowing us to assess the effectiveness of the proposed framework comprehensively. More details are shown below.

\subsection{NLVR2}
The NLVR2~\cite{nlvr2} dataset is curated to advance research in computer vision and natural language processing for visual reasoning tasks. Its main objective is to enable models to determine if two images share common objects or scenes using provided natural language descriptions. With 107,292 examples of human-written English sentences grounded in pairs of photographs, NLVR2 offers linguistic diversity and visually complex images. The dataset is divided into subsets: the training set contains 86,373 examples, the development set consists of 6,982 examples, Test-P comprises 6,967 examples, and Test-U includes 6,970 examples. The primary evaluation metric is Accuracy (Acc), reflecting the proportion of correctly predicted image pairs. These evaluation metrics aid researchers in assessing model performance, facilitating comparisons and guiding improvements.

\subsection{COCO}
The COCO~\cite{Imagecaption_2014} dataset is a valuable resource for both image-text retrieval and image caption tasks, containing a vast amount of annotated data. It includes 82,783 training images with 413,915 captions, 40,504 validation images with 202,520 captions, and 40,775 testing images with 379,249 captions. 
For the image-text retrieval task, Recall@k serves as a useful evaluation metric. It quantifies the proportion of relevant results that are correctly retrieved within the top-k ranked items. This metric is valuable for assessing the model's ability to recall relevant captions when given an image query and vice versa.
For the image caption task, evaluation metrics such as CIDEr and SPICE are commonly used. CIDEr (Consensus-based Image Description Evaluation) leverages consensus-based scoring by comparing generated captions to multiple reference captions, providing a measure of the quality of the generated captions. SPICE (Semantic Propositional Image Caption Evaluation) considers the semantic structure of the captions by evaluating their ability to describe the image content accurately.

\subsection{Flickr30k}
The Flickr30k~\cite{flick30k} dataset is widely utilized for image caption and image-text retrieval tasks, providing a substantial collection of images with associated captions. It consists of three distinct subsets: a training set comprising 29,000 images and 145,000 captions, a validation set containing 1,000 images and 5,000 captions, and a test set with 1,000 images and 5,000 captions. This dataset provides researchers with a diverse range of images and associated textual descriptions, enabling the development and evaluation of models for various image understanding tasks. In the experiments of this paper, we focus on evaluating the performance of the MADTP compressed models for the image-text retrieval task using the Flickr30k dataset. To ensure consistency with evaluation practices used in the COCO~\cite{Imagecaption_2014} dataset, we employed the same Recall@k metric as the final evaluation metric.

\subsection{VQA 2.0}
The VQA 2.0~\cite{vqa} dataset serves as a widely adopted resource for Visual Question Answering (VQA) task, where models are tasked with answering questions related to images. It is an extended version of the original VQA dataset, addressing its limitations and providing a more comprehensive evaluation setup. The dataset is derived from the COCO~\cite{Imagecaption_2014} dataset and is divided into three main subsets: training, validation, and testing. The training set consists of approximately 82,783 images with 443,757 associated questions. The validation set contains around 40,504 images with 214,354 questions, while the testing set comprises about 81,434 images with 447,793 questions. Notably, the testing set is further divided into two distinct subsets: test-dev and test-std. The test-dev subset is designated for model development and fine-tuning purposes, while the test-std subset is reserved for official evaluation and facilitates performance comparisons. Evaluation of models on the VQA 2.0 dataset employs various metrics. The primary metric is Accuracy (Acc), which measures the proportion of correctly answered questions. Additionally, the dataset provides per-question-type and per-answer-type accuracy metrics, allowing for a more detailed analysis of model performance across different question and answer categories. 

\subsection{GFLOPs}

GFLOPs (Giga Floating Point Operations per Second) is a widely adopted metric for quantifying the computational costs of computer systems, particularly in the fields of deep learning and artificial intelligence. It measures the number of floating-point operations that a system can perform in one second, with "Giga" representing one billion ($10^{9}$) operations. 
In this paper, the GFLOPs can vary for different inputs due to the instance-level dynamic pruning scheme employed by our MADTP. Therefore, in our experiments, we opted to calculate the averaged GFLOPs over the entire dataset to effectively measure the computational overhead of the compressed model.

\section{Implementation details}
\label{sec:implementation}

In our experiments, we employ the MADTP framework to compress Vision-Language Transformers, specifically the CLIP~\cite{CLIP} and BLIP~\cite{li2022blip} models. These models are initialized with pretrained weights obtained from the official implementation of \cite{2023Upop}.
Table~\ref{tab:hp_clip} and Table~\ref{tab:hp_blip} present detailed hyperparameter settings for each model during the compression training process. Further, Table~\ref{table structure hyperparameters} details the architecture configures of the Vision-Language Transformers used in different multimodal models.
In our experimental setup, we train the models using 8 A100 GPUs, with a fixed batch size of 32. Note that, unlike the two-stage approach employed in Upop~\cite{2023Upop}, our method is a one-stage approach that eliminates the search stage, resulting in a significant reduction in training time. The MADTP framework exhibits fast convergence, often achieving promising results within just 1-2 epochs. For example, in the case of BLIP-VQA, impressive performance is observed after only 3 epochs of training.
In terms of specific hyperparameters, the number of learnable tokens is consistently set to 100, and the channel dimension is set to 768 across different models. Additionally, the hyperparameter $\alpha$ in the loss function is consistently set to $0.1$. To enable parallel training, we incorporated the "max-keep" operation within each mini-batch to retain crucial tokens.
We will release the code, allowing others to build upon our work.

\begin{table}[ht]
    \small
    \centering
    \vspace{-2mm}
  \begin{tabular}{l @{\hspace{1.0\tabcolsep}} c @{\hspace{1.0\tabcolsep}} c}
    \toprule
    \multirow{2}{*}{Hyperparameters} & \multicolumn{2}{c}{CLIP \cite{CLIP}} \\ 
    \cmidrule{2-3}
    & \multirow{1}{*}{\makecell{COCO \cite{Imagecaption_2014}}} & \multirow{1}{*}{\makecell{Flickr30K \cite{flick30k}}} \\
    \midrule
    Optimizer & \multicolumn{2}{c}{AdamW \cite{loshchilov2017decoupled}}  \\
    AdamW $\beta$   & \multicolumn{2}{c}{(0.9, 0.999)} \\
    Weight decay & \multicolumn{2}{c}{0.2} \\
    Batch size & \multicolumn{2}{c}{32} \\
    Train epochs & 5 & 10\\
    Train LR & \multicolumn{2}{c}{1e-5}\\
    Learnable token numbers $K$  & \multicolumn{2}{c}{100} \\
    Learnable token dimensions $d_k$ &  \multicolumn{2}{c}{768} \\
    Loss weight $\alpha$ & \multicolumn{2}{c}{0.1}  \\
    Prune operation &  \multicolumn{2}{c}{max-keep} \\
    Train LR schedule & \multicolumn{2}{c}{CosineLRScheduler \cite{loshchilov2016sgdr}} \\
    Data augmentation & \multicolumn{2}{c}{\makecell{RandomAugment \cite{cubuk2020randaugment}}} \\
    \bottomrule
  \end{tabular}
    \vspace{-2mm}
  \caption{Training hyperparameters for compressing CLIP-based models on both COCO and Flickr30K datasets.}
   \vspace{-4mm}
\label{tab:hp_clip}
\end{table}

\section{Supplementary Experiments and Analyses}
\label{sec:experiments}
\subsection{Comparison with Token Pruning}
In this study, we conduct a comparative analysis between our MADTP and some recent token pruning techniques, including CrossGET~\cite{shi2023crossget} and ELIP~\cite{guo2023elip}.
However, it should be noted that these methods have not been formally published and are currently only available on the arXiv website. Hence, we do not include them in our main paper. 

Detailed comparisons are shown in Table~\ref{tab:clip_crossget} and Table~\ref{tab:blip_elip}.
Specifically, CrossGET~\cite{shi2023crossget} introduces the use of cross tokens as guidance for both modalities and employs the single-modality token merge method~\cite{Bolya_Fu_Dai_Zhang_Feichtenhofer_Hoffman} for accelerating VLTs. On the other hand, ELIP~\cite{guo2023elip} proposes a vision token pruning and merging method that removes less influential tokens based on the supervision of language outputs.
Both of these methods overlook the significance of modality alignment guidance in the multimodal token pruning process. Additionally, they belong to the category of static token pruning, which cannot achieve adaptive dynamic compression for Vision-Language Transformers. In contrast, our MADTP method introduces the Multi-modality Alignment Guidance (MAG) module, which enables modality alignment guidance during VLT compression. Further, we design the Dynamic Token Pruning (DTP) module, which can achieve both input instance- and layer-wise compression of VLTs.
Due to the differences in experimental settings and challenges related to code release, we focus on comparing the final compression results with these two methods.
The experimental results clearly show that our MADTP achieves superior compression performance compared to CrossGET~\cite{shi2023crossget} and ELIP~\cite{guo2023elip}, which provide strong evidence for the effectiveness of our approach.

\begin{table*}[!htp]
    \small
    \centering
    \setlength{\tabcolsep}{3.5mm}{
  \begin{tabular}{l @{\hspace{1.0\tabcolsep}} c @{\hspace{1.0\tabcolsep}} c c @{\hspace{1.5\tabcolsep}} c @{\hspace{1.5\tabcolsep}} c}
    \toprule
    \multirow{4}{*}{Hyperparameters} & \multirow{2}{*}{\makecell{BLIP-NLVR \\ \cite{li2022blip}}} & \multirow{2}{*}{\makecell{BLIP-Caption\\ \cite{li2022blip}}} & \multirow{2}{*}{\makecell{BLIP-VQA\\ \cite{li2022blip}}} & \multicolumn{2}{c}{\multirow{2}{*}{\makecell{BLIP-Retrieval\\ \cite{li2022blip}}}}  \\ 
    & & & & & \\
    \cmidrule{2-6}
    & \multirow{2}{*}{\makecell{NLVR2\\ \cite{nlvr2}}} & \multirow{2}{*}{\makecell{COCO\\ \cite{Imagecaption_2014}}} & \multirow{2}{*}{\makecell{VQAv2\\ \cite{vqa}}} & \multirow{2}{*}{\makecell{COCO\\ \cite{Imagecaption_2014}}} & \multirow{2}{*}{\makecell{Flickr30K\\ \cite{flick30k}}} \\
    & & & & & \\
    \midrule
    Optimizer & \multicolumn{5}{c}{AdamW \cite{loshchilov2017decoupled}} \\
    AdamW $\beta$ & \multicolumn{5}{c}{(0.9, 0.999)} \\
    Weight decay & \multicolumn{5}{c}{0.05} \\
    Batch size & \multicolumn{5}{c}{32} \\
    Train epochs & 15 & 5 & 3 & 5 & 10 \\
    Train LR & 3e-6 & 1e-5 & 2e-5 & 1e-6 & 1e-5 \\
    Learnable token numbers $K$  & \multicolumn{5}{c}{100} \\
    Learnable token dimensions $d_k$ &  \multicolumn{5}{c}{768} \\
    Loss weight $\alpha$ & \multicolumn{5}{c}{0.1}  \\
    Prune operation &  \multicolumn{5}{c}{max-keep} \\
    Train LR schedule & \multicolumn{5}{c}{CosineLRScheduler \cite{loshchilov2016sgdr}} \\
    Data augmentation & \multicolumn{5}{c}{RandomAugment \cite{cubuk2020randaugment}}\\
    \bottomrule
  \end{tabular}
  }
  \vspace{-1mm}
\caption{Training hyperparameters for compressing BLIP-based models on five kinds of datasets.}
\vspace{-2mm}
\label{tab:hp_blip}
\end{table*}

\begin{table*}[!htp]
    \small
    \centering  
  \begin{tabular}{lccccccccc}
    \toprule
    \multirow{2}{*}{Model} & \multirow{2}{*}{\makecell{Input\\resolution}} & \multicolumn{4}{c}{Vision Transformer} & \multicolumn{4}{c}{Language Transformer} \\
    & & number & layers & width & heads & number & layers & width & heads \\
    \midrule
    BLIP-NLVR \cite{li2022blip}& 384$\times$384 & 2\textsuperscript{$*$} & 12 & 768 & 12 & 1 & 12 & 768 & 12 \\
    BLIP-Caption \cite{li2022blip}& 384$\times$384 & 1 & 12 & 768 & 12 & 1 & 12 & 768 & 12 \\
    BLIP-VQA \cite{li2022blip}& 480$\times$480 & 1 & 12 & 768 & 12 & 2 & 12 & 768 & 12 \\
    BLIP-Retrieval \cite{li2022blip} & 384$\times$384 & 2 & 12 & 768 & 12 & 2 & 12 & 768 & 12 \\
    CLIP \cite{CLIP} & 336$\times$336 & 2 & 24 & 1024 & 16 & 2 & 12 & 768 & 12 \\
    \bottomrule
  \end{tabular}
 \vspace{-1mm}
  \caption{Architecture configures of all models used in our experiments. The superscript \textsuperscript{$*$} indicates 2 Transformers share parameters. 
  }
  \vspace{-2mm}
\label{table structure hyperparameters}
\end{table*}

\begin{table*}[!htp]
\small
\centering
\setlength{\tabcolsep}{3.8mm}{
\begin{tabular}{c|ccc|ccc|c}
\toprule
&  \multicolumn{3}{c|}{Image$\to$Text} & \multicolumn{3}{c|}{Text$\to$Image} &  \\ 
\multirow{-2}{*}{Approach} & \multicolumn{1}{c}{R@1}   & \multicolumn{1}{c}{R@5}   & \multicolumn{1}{c|}{R@10} & \multicolumn{1}{c}{R@1}   & \multicolumn{1}{c}{R@5}   & \multicolumn{1}{c|}{R@10}& \multirow{-2}{*}{GFLOPs}   \\ \midrule
                            ToMe\textdaggerdbl ~\cite{shi2023crossget} & \multicolumn{1}{c}{90.8}& \multicolumn{1}{c}{99.2}& \multicolumn{1}{c|}{99.5}& \multicolumn{1}{c}{78.1}& \multicolumn{1}{c}{95.3}& \multicolumn{1}{c|}{97.7}& - \\
                            CrossGET~\cite{shi2023crossget} & \multicolumn{1}{c}{92.1}& \multicolumn{1}{c}{\textbf{99.7}}& \multicolumn{1}{c|}{99.8}& \multicolumn{1}{c}{79.6}& \multicolumn{1}{c}{\textbf{97.5}}& \multicolumn{1}{c|}{98.0}& - \\
                            UPop~\cite{2023Upop} & \multicolumn{1}{c}{93.2}& \multicolumn{1}{c}{99.4}& \multicolumn{1}{c|}{99.8}& \multicolumn{1}{c}{80.5}& \multicolumn{1}{c}{95.4}& \multicolumn{1}{c|}{97.6}& 201.1 \\
                            \textbf{MADTP (Ours)} & \multicolumn{1}{c}{\textbf{93.9}}& \multicolumn{1}{c}{99.5}& \multicolumn{1}{c|}{\textbf{99.8}}& \multicolumn{1}{c}{\textbf{83.3}} & \multicolumn{1}{c}{97.0} & \multicolumn{1}{c|}{\textbf{98.5}}& 178.8 \\              
\bottomrule
\end{tabular}
}
\vspace{-1mm}
\caption{Performance comparisons of different methods when compressing CLIP on the Flickr30K dataset of the Image-Text Retrieval task. The R@1, R@5, and R@10 are the higher the better. The best results are in bold. The symbol \textdaggerdbl \ represents the model implementation is derived from CrossGET~\cite{shi2023crossget}.}
\vspace{-2mm}
\label{tab:clip_crossget}
\end{table*}

\begin{table*}[!htp]
\small
\centering
\setlength{\tabcolsep}{1.4mm}{
\begin{tabular}{c|ccc|ccc|c|ccc|ccc|c}
\toprule
& \multicolumn{7}{c|}{Flickr30K} & \multicolumn{7}{c}{COCO} \\ \cline{2-15} 
& \multicolumn{3}{c|}{Image$\to$Text}& \multicolumn{3}{c|}{Text$\to$Image} & & \multicolumn{3}{c|}{Image$\to$Text}& \multicolumn{3}{c|}{Text$\to$Image}& \\ 
\multirow{-3}{*}{Approach} & \multicolumn{1}{c}{R@1}   & \multicolumn{1}{c}{R@5}   & \multicolumn{1}{c|}{R@10} & \multicolumn{1}{c}{R@1}   & \multicolumn{1}{c}{R@5}   & \multicolumn{1}{c|}{R@10} & \multirow{-2}{*}{GFLOPs}  & \multicolumn{1}{c}{R@1}   & \multicolumn{1}{c}{R@5}   & \multicolumn{1}{c|}{R@10} & \multicolumn{1}{c}{R@1}   & \multicolumn{1}{c}{R@5}   & \multicolumn{1}{c|}{R@10} & \multirow{-2}{*}{GFLOPs}  
\\ \midrule
                             
                             EViT\textdagger~\cite{guo2023elip} & \multicolumn{1}{c}{87.3}& \multicolumn{1}{c}{98.5}  & \multicolumn{1}{c|}{99.4}& \multicolumn{1}{c}{75.1}& \multicolumn{1}{c}{93.5}& \multicolumn{1}{c|}{96.4}& 48.0 & \multicolumn{1}{c}{66.8}& \multicolumn{1}{c}{88.9}& \multicolumn{1}{c|}{93.9}& \multicolumn{1}{c}{50.8}& \multicolumn{1}{c}{77.9}& \multicolumn{1}{c|}{86.3}& 48.0 \\ 
                             ToMe\textdagger~\cite{guo2023elip} & \multicolumn{1}{c}{91.5}& \multicolumn{1}{c}{98.8}  & \multicolumn{1}{c|}{99.4}& \multicolumn{1}{c}{80.5}& \multicolumn{1}{c}{95.6}& \multicolumn{1}{c|}{97.9}& 69.8 & \multicolumn{1}{c}{71.5}& \multicolumn{1}{c}{91.6}& \multicolumn{1}{c|}{95.9}& \multicolumn{1}{c}{55.3}& \multicolumn{1}{c}{81.2}& \multicolumn{1}{c|}{88.7}& 69.8 \\ 
                             ELIP~\cite{guo2023elip} & \multicolumn{1}{c}{92.2}& \multicolumn{1}{c}{99.1}  & \multicolumn{1}{c|}{99.7}& \multicolumn{1}{c}{80.3}& \multicolumn{1}{c}{96.0}& \multicolumn{1}{c|}{98.0}& 93.4 & \multicolumn{1}{c}{72.0}& \multicolumn{1}{c}{91.9}& \multicolumn{1}{c|}{95.9}& \multicolumn{1}{c}{56.3}& \multicolumn{1}{c}{81.2}& \multicolumn{1}{c|}{88.7}& 93.4 \\ 
                             UPop~\cite{2023Upop} & \multicolumn{1}{c}{94.0}& \multicolumn{1}{c}{99.5}  & \multicolumn{1}{c|}{99.7}& \multicolumn{1}{c}{82.0}& \multicolumn{1}{c}{95.8}& \multicolumn{1}{c|}{97.6}& 91.0 & \multicolumn{1}{c}{77.4}& \multicolumn{1}{c}{93.4}& \multicolumn{1}{c|}{97.0}& \multicolumn{1}{c}{59.8}& \multicolumn{1}{c}{83.1}& \multicolumn{1}{c|}{89.8}& 88.3 \\ 
                             \textbf{MADTP (Ours)}& \multicolumn{1}{c}{\textbf{95.1}}  & \multicolumn{1}{c}{\textbf{99.5}}& \multicolumn{1}{c|}{\textbf{99.7}}  & \multicolumn{1}{c}{\textbf{82.3}} & \multicolumn{1}{c}{\textbf{96.2}} & \multicolumn{1}{c|}{\textbf{98.0}} & 74.5 & \multicolumn{1}{c}{\textbf{79.1}} & \multicolumn{1}{c}{\textbf{94.2}} & \multicolumn{1}{c|}{\textbf{97.2}} & \multicolumn{1}{c}{\textbf{60.3}} & \multicolumn{1}{c}{\textbf{83.6}} & \multicolumn{1}{c|}{\textbf{89.9}}  & 87.4
                             \\ \bottomrule
\end{tabular}
}
\vspace{-1mm}
\caption{Performance comparisons of different methods when compressing BLIP on the Flickr30K and COCO datasets of the Image-Text Retrieval task. The R@1, R@5, and R@10 are the higher the better. The best results are in bold. The symbol \textdagger \ represents the model implementation is derived from ELIP~\cite{guo2023elip}.}
\vspace{-2mm}
\label{tab:blip_elip}
\end{table*}

\begin{table*}[!htp]
\small
\centering 
\setlength{\tabcolsep}{2.5mm}{
\begin{tabular}{c|ccccll}
\toprule
Approach & \makecell{Reduce ratio \\ (Params)} & \makecell{Reduce ratio\\ (GFLOPs)} & Dev Acc & Test Acc & Params & GFLOPs \\ \midrule
Uncompressed    & - & - & 82.48   & 83.08  & 259.45  & 132.54  \\ \midrule
Parameter pruning~\cite{2023Upop}  & 0.15 & - & 81.54   & 82.35  & \textbf{219}  & 117.32    \\
\multirow{1}{*}{Parameter pruning~\cite{2023Upop} + \textbf{MADTP}} & 0.15   & 0.3   & \textbf{81.80}  & \textbf{82.52}   & 220{\scriptsize\color[HTML]{68CBD0} $\uparrow$0.4\%}    & \textbf{92.75} {\scriptsize\color[HTML]{FE0000} $\downarrow$20.8\%}  \\ 
\bottomrule
\end{tabular}
}
\vspace{-2mm}
\caption{The orthogonality of our MADTP framework with parameter pruning techniques. Compress BLIP on the NLVR2 dataset for visual reasoning task. Reduce ratio (Params) represents the proportion of model parameter compression, and Reduce ratio (GFLOPs) denotes the compression ratio of model computational costs. The experimental results demonstrate that combining our approach with parameter pruning techniques yields superior compression performance.}
\vspace{-4mm}
\label{tab:Orthogonality}
\end{table*}

\begin{table}[!htp]
\small
\centering 
\setlength{\tabcolsep}{0.5mm}{
\begin{tabular}{c|c|ccc}
\toprule
Approach & Modality & Dev Acc & Test Acc & GFLOPs \\ \midrule
Uncompressed & - & 82.48   & 83.08  & 132.54  \\ \midrule
& vision only  & 80.04   & 80.50  & 67.69  \\
& language only  & 74.67   & 75.01  & 129.54  \\
\multirow{-3}{*}{STP} & vision and language  & 78.08   & 77.61  & 68.31  \\ \midrule
& vision only  & \textbf{82.27}   & 82.45  & 66.41  \\
& language only  & 77.33   & 77.58  & 128.98  \\
\multirow{-3}{*}{\textbf{MADTP}} & vision and language  & 81.97   & \textbf{82.85}  & \textbf{66.16}  \\
\bottomrule
\end{tabular}
}
\vspace{-2mm}
\caption{Ablation studies of MADTP on different modalities.}
\vspace{-6mm}
\label{tab:different_modality}
\end{table}

\subsection{Orthogonality with Parameter Pruning}
In this section, we conduct experiments to validate the orthogonality of our MADTP framework with parameter pruning techniques. The detailed results are presented in Table~\ref{tab:Orthogonality}. 
Here are the specifics of the experimental setup: we firstly apply a parameter pruning approach~\cite{2023Upop} to the BLIP model, using a compression ratio of 0.15 on the NLVR2 dataset as the initial compression step. Subsequently, we further accelerate the compressed model using our MADTP with a reduce ratio of 0.3. 
The objective of this additional pruning step is to dynamically eliminate non-critical tokens, thereby further enhancing model efficiency.
The thorough experimental results confirm the orthogonality of our MADTP framework with parameter pruning approaches. 
In detail, after applying our MADTP method, the model exhibits a 0.26\% increase in accuracy on the dev set and a 0.17\% increase on the test set. The GFLOPs of the compressed model decrease by 20.8\%, indicating a substantial reduction in computational costs. Remarkably, despite these improvements, the model's parameters only increases by a mere 0.4\%. 
Therefore, combining both pruning schemes in a joint compression strategy yields outstanding compression results. Our future work involves integrating a parameter pruning scheme into the proposed MADTP framework for comprehensive VLT compression.

\subsection{Compression on different modalities}
We also perform ablation studies on applying the proposed MADTP method to compress different modalities for VLTs, and the detailed results can be found in Table~\ref{tab:different_modality}. 
Due to the varying importance of different modalities in accomplishing the final task and the different computational costs associated with each modality branch, individually compressing different modalities has a significant impact on the overall performance of the compressed model.
In our experiment, we separately compressed various modal branches of the BLIP model on the NLVR2 dataset, including the only vision branch, only language branch, and the combined vision and language branch. 
The experimental results indicate that the visual branch has higher token redundancy, allowing for significant reductions in computational costs through token pruning.
Conversely, the text branch has lower computational cost and is essential for multimodal tasks. Thus, compressing the text branch has a more substantial impact on model performance, albeit with minimal decrease in GFLOPs.
These observations aligns with the finding of the CrossGET~\cite{shi2023crossget} method.
However, our MADTP method additionally accounts for modality alignment and integrates an adaptive token pruning mechanism, facilitating collaborative compression of both modalities and achieving superior compression results.

\begin{table}[!htp]
\small
\centering
\setlength{\tabcolsep}{2mm}{
\begin{tabular}{cl|ccc}
\toprule
\multicolumn{2}{c|}{Components of MADTP}  & Dev Acc & Test Acc & GFLOPs \\ \midrule
\multirow{7}{*}{TIS} & only w/$S_{\text{self}}$  & 81.49   & 82.13    & 70.46  \\ 
                     & only w/$S_{\text{token}}$ & 80.68   & 81.00    & 66.74  \\ 
                     & only w/$S_{\text{cls}}$   & 81.62   & 82.25    & 69.67  \\  
                     & $S_{\text{self}}$ \& $S_{\text{token}}$      & 81.79   & 82.32    & 67.08  \\ 
                     & $S_{\text{self}}$ \& $S_{\text{cls}}$      & 81.40   & 82.35    & 70.67  \\
                     & $S_{\text{token}}$ \& $S_{\text{cls}}$      & 81.76   & 82.41    & 66.19  \\  
                     & $S_{\text{self}}$ \& $S_{\text{token}}$ \& $S_{\text{cls}}$ & \textbf{81.97}   & \textbf{82.85}    &  \textbf{66.16}  \\ \bottomrule
\end{tabular}
}
\vspace{-1mm}
\caption{Results of compressing the BLIP model on the NLVR2 dataset with different token importance scores.}
\vspace{-1mm}
\label{tab:TIS}
\end{table}

\begin{table}[!htp]
\small
\setlength{\tabcolsep}{1.8mm}{
\begin{tabular}{c|cccc}
\toprule
Setting & Batch size & Temperature & Test Acc & GFLOPs \\ \midrule
Baseline & 16 & 1.26 & 82.35 & 67.62  \\   \midrule
\multirow{6}{*}{Inference} & \multirow{2}{*}{1}  & 1.26  & 77.90  & 38.46 \\  
& & 0.44  & 81.86   & 67.04  \\ \cline{2-5}
& \multirow{2}{*}{4}  & 1.26  & 81.04  & 52.13  \\ 
& & 0.89  & 82.20  & 66.97  \\ \cline{2-5}
& \multirow{2}{*}{32} & 1.26  & \textbf{82.36}   & 75.08  \\  
& & 1.43  & 82.08   & 68.37  \\ \bottomrule
\end{tabular}
}
\vspace{-1mm}
\caption{
The performance of the 0.5 compressed BLIP model on NLVR2 dataset when using different batch sizes during inference. Our baseline model is trained with a batch size of 16, and the GFLOPs with different batch sizes can be adjusted by controlling the temperature to maintain consistency with the baseline.
}
\vspace{-1mm}
\label{tab:batchsize}
\end{table}

\begin{table}[!htp]
\small
\setlength{\tabcolsep}{2.8mm}{
\begin{tabular}{c|c|ccc}
\toprule
Batch size & Sorted & Dev Acc & Test Acc & GFLOPs   \\ \midrule
\multirow{2}{*}{1}   & N    & 76.96   & 77.90  & 38.46    \\ 
& Y  & -  & {\color[HTML]{68CBD0} 77.74$\downarrow$} & 38.50 \\ \midrule
\multirow{2}{*}{4}  & N  & 80.48    & 81.04   & 52.13   \\ 
& Y    & -    & {\color[HTML]{FE0000} 81.16$\uparrow$} & 53.36 \\ \midrule
\multirow{2}{*}{16} & N & 81.64   & 82.35   & 67.62   \\ 
& Y  & -   & {\color[HTML]{FE0000} 82.59$\uparrow$} & 67.61 \\ \midrule
\multirow{2}{*}{32} & N & 81.96  & 82.36   & 75.08  \\ 
& Y   & - & {\color[HTML]{FE0000} 82.74$\uparrow$} & 73.78 \\ \bottomrule
\end{tabular}
}
\vspace{-1mm}
\caption{Performance of the 0.5 compressed BLIP model on the NLVR2 dataset when using different instance order. Sorted Y means we first sort the instances according to their difficulty and then use the compressed model for inference.}
\vspace{-2mm}
\label{tab:sorted}
\end{table}

\begin{figure*}[!htp]
  \centering
  \includegraphics[width=0.99\linewidth]{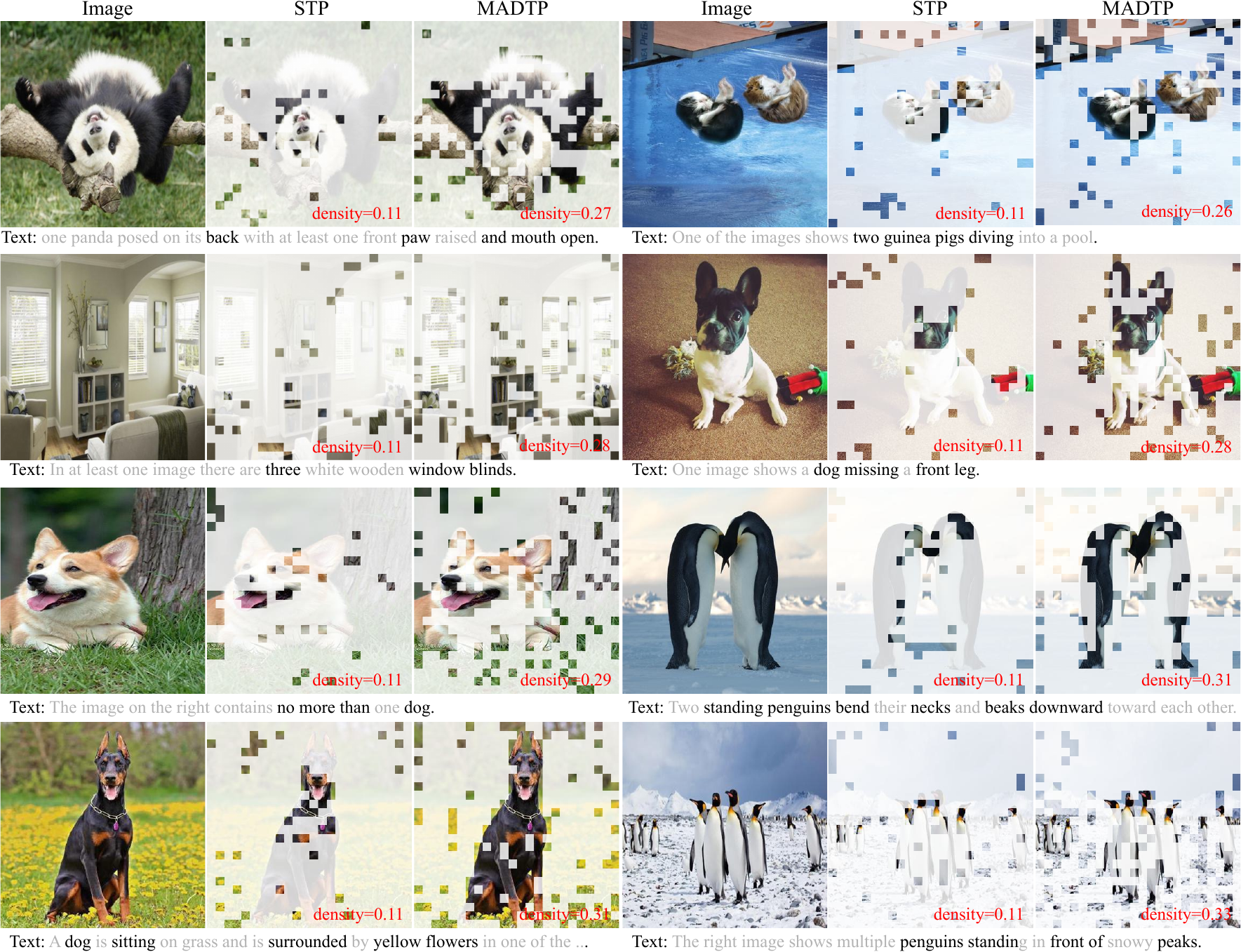}
  \vspace{-1mm}
  \caption{Visualization comparisons of token pruning results between STP and MADTP, providing strong evidence that our approach emphasizes modality correlation, effectively avoids pruning crucial tokens and dynamically adjusts pruning ratio according to inputs.
  }
  \label{fig:compare_stp}
  \vspace{-5mm}
\end{figure*}

\begin{figure}[!htp]
  \centering
  \includegraphics[width=0.95\linewidth]{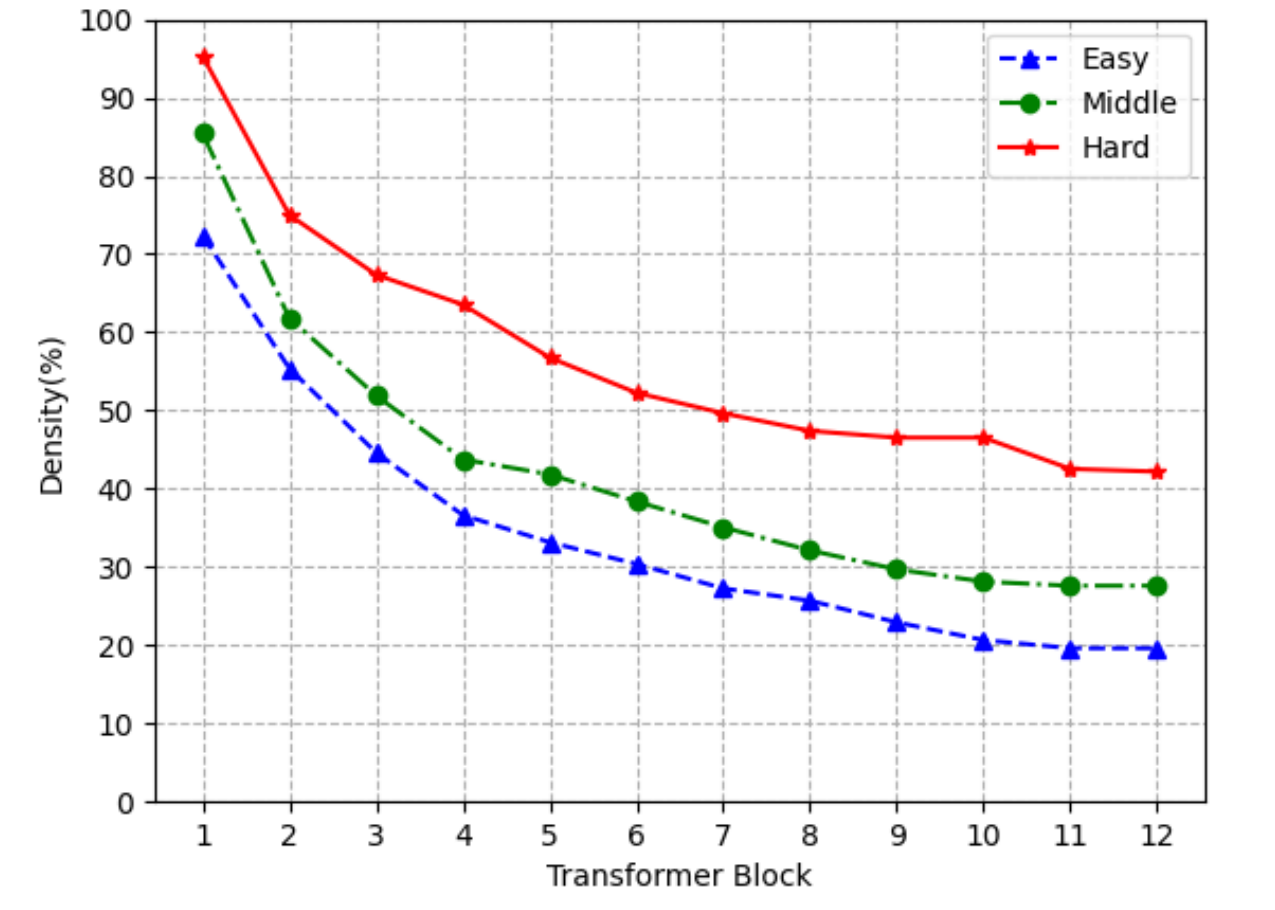}
  \vspace{-1mm}
  \caption{Comparisons of MADTP token pruning in each transformer block for samples of different instance complexity levels, including Easy, Middle, and Hard samples. The density represents the ratio of retained tokens to the total number of original tokens.}
  \vspace{-7mm}
  \label{fig:reduce_density}
\end{figure}

\begin{figure*}[!htp]
  \centering
  \includegraphics[width=\linewidth]{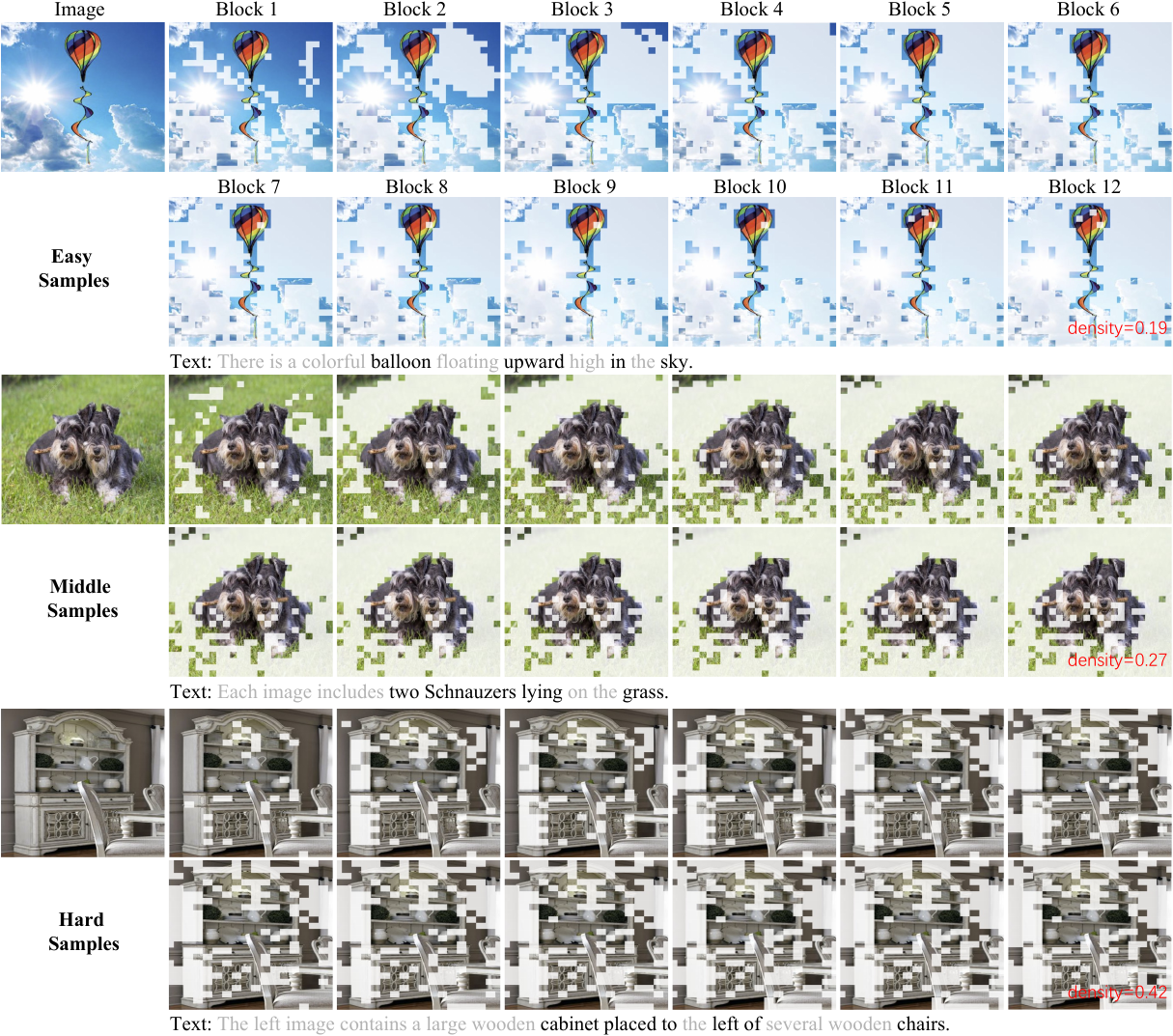}
  \caption{Visualization of the compressed results of MADTP on samples with different levels of instances complexity, including Easy, Middle, and Hard samples. The density represents the ratio of retained tokens to the total number of original tokens.
  }
  \label{fig:diffrent_sample}
\end{figure*}

\subsection{Effect of Hyperparameters}
In this section, we conduct additional ablation studies to validate the hyperparameters that affect the performance of MADTP.
Firstly, we extend our analysis about the Token Importance Scores($\text{TIS}$), as shown in Table~\ref{tab:TIS}.
Furthermore, we discover that our MADTP is significantly influenced by the batch size during the inference stage, as demonstrated in Table~\ref{tab:batchsize}.
The reason behind this observation is that we adopt the max-keep pruning strategy in the token pruning process, which selects the maximum number of tokens to be retained across input instances in a mini-batch.
Therefore, when using a smaller batch size for model inference, the GFLOPs significantly decrease, leading to a decline in performance.
However, by adjusting the temperature parameter $T$, we can increase the GFLOPs with the smaller batch size to match the baseline model, thereby restoring the performance.
This experiment proves the strong correlation between the compressed model's performance and GFLOPs. 
In addition, as shown in Table~\ref{tab:sorted}, we observe that sorting the input instances based on their difficulty during inference leads to improved performance. This finding suggests that applying the max-keep strategy to sorted input instances can further enhance compressed models' performance. 

\subsection{Visualization of MADTP}
\begin{figure*}[!htp]
  \centering
  \vspace{-5mm}
  \includegraphics[width=0.95\linewidth]{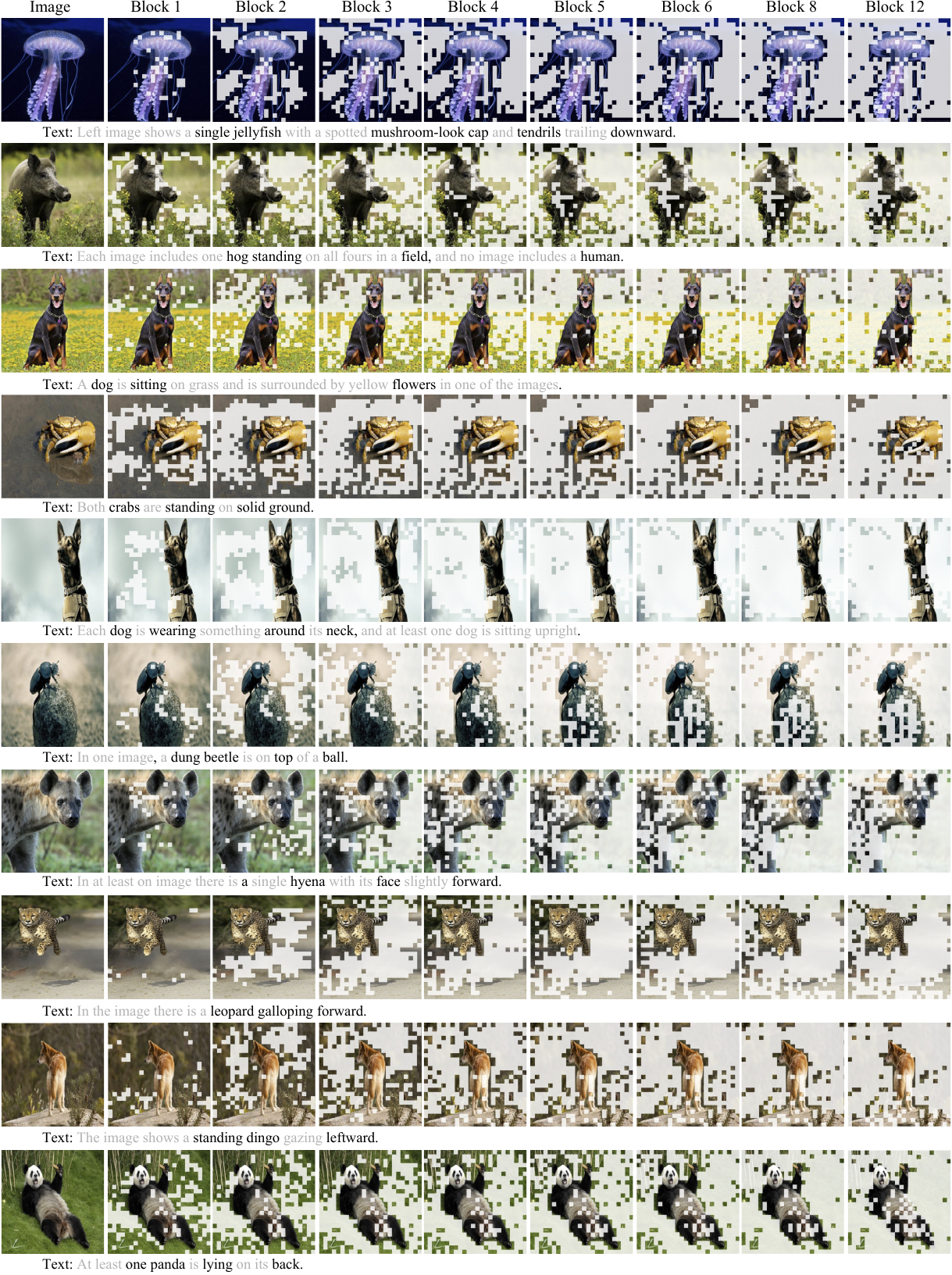}
  \vspace{-3mm}
  \caption{Visualization of our MADTP’s compressed BLIP results on \textbf{Easy Samples} from the NLVR2 dataset at each transformer block.
  }
  \label{fig:easy samples}
  \vspace{-2mm}
\end{figure*}

\begin{figure*}[!htp]
  \centering
  \vspace{-5mm}
  \includegraphics[width=0.95\linewidth]{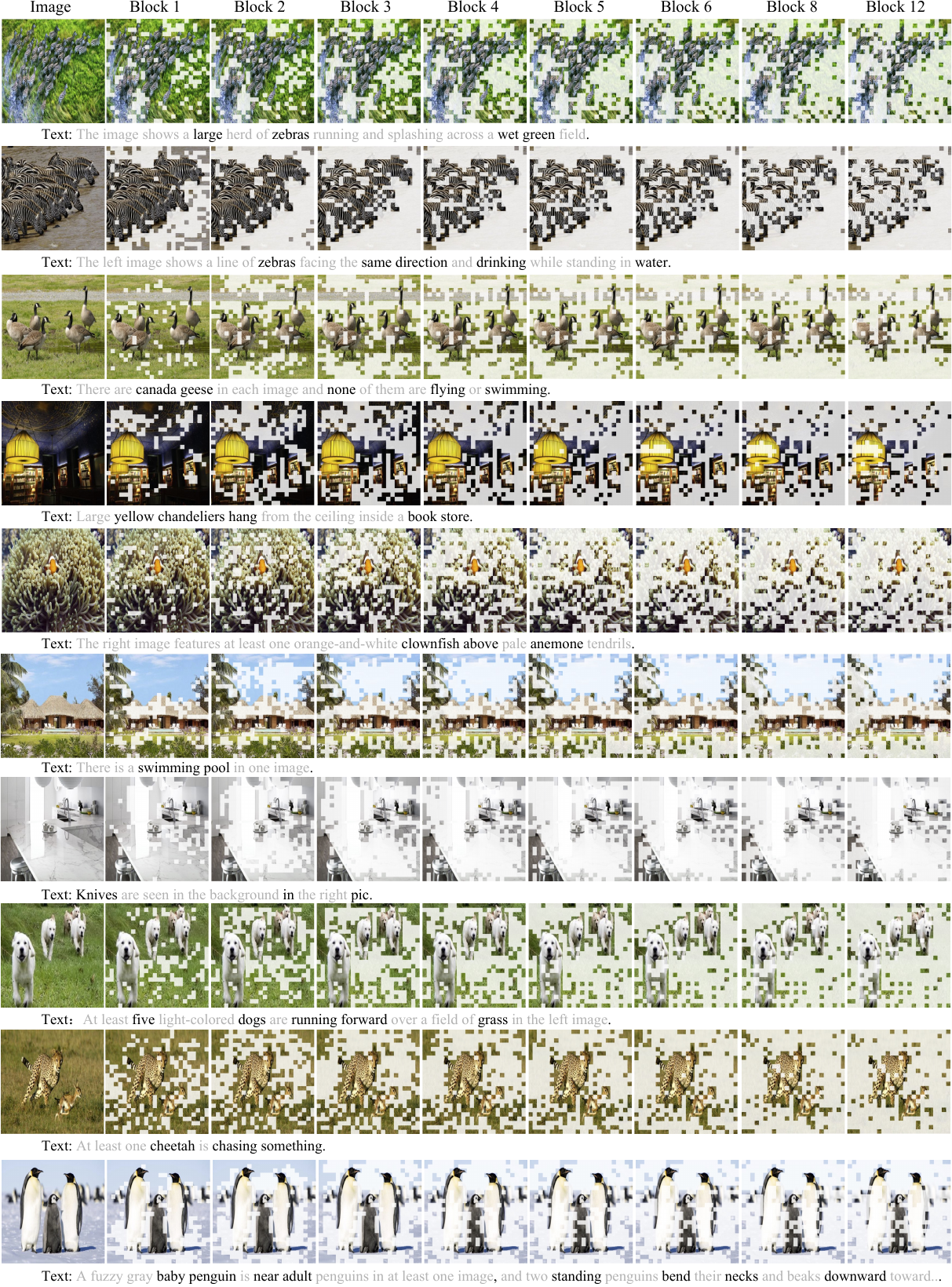}
  \caption{
  Visualization of our MADTP’s compressed BLIP results on \textbf{Hard Samples} from the NLVR2 dataset at each transformer block.
  }
  \label{fig:hard samples}
  \vspace{-2mm}
\end{figure*}

In this section, we visualize the token pruning results of the proposed MADTP framework using a compressed BLIP model with a reduction ratio of 0.5 on the NLVR2 dataset.
In Fig.~\ref{fig:compare_stp}, we present an extended visualization comparison between Static Token Pruning (STP) and our MADTP approach. It is evident that our MADTP emphasizes the correlation between modalities and successfully avoids pruning critical tokens.
Additionally, we further visualize MADTP token pruning in each transformer block for samples with different instance complexity levels, including Easy, Middle, and Hard samples. Fig.~\ref{fig:reduce_density} illustrates the token density in the visual branch of VLTs at each transformer block, while Fig.~\ref{fig:diffrent_sample} showcases the specific positions of token pruning in each block.
These visualizations demonstrate the adaptive dynamic compression capability of the proposed MADTP framework for different input instances. 
Finally, we show additional visualizations of token compression using the MADTP framework for easy and hard samples in Fig.~\ref{fig:easy samples} and Fig.~\ref{fig:hard samples}. These visualizations further validate the effectiveness of MADTP in dynamically compressing tokens for Vision-Language Transformers.

\clearpage

\end{document}